\documentclass[twoside,11pt]{article}
\usepackage{blindtext}

\usepackage{amsmath,amsfonts}
\usepackage{array}
\usepackage[caption=false,font=normalsize,labelfont=sf,textfont=sf]{subfig}
\usepackage{textcomp}
\usepackage{stfloats}
\usepackage{url}
\usepackage{verbatim}
\usepackage{booktabs}
\usepackage{graphicx}
\usepackage{enumitem}
\usepackage{bm}
\usepackage{braket}
\usepackage{xcolor}
\usepackage[ruled,vlined]{algorithm2e}

\usepackage[preprint]{jmlr2e}
\usepackage{multirow} % Include this package for multirow

\DeclareMathOperator*{\argmax}{arg\,max}
\DeclareMathOperator*{\argmin}{arg\,min}

\usepackage{lastpage}
\jmlrheading{xx}{2026}{1-\pageref{LastPage}}{x/xx; Revised x/xx}{x/xx}{xx-xxxx}{Luis A. Ortega, Simón Rodriguez-Santana and Daniel Hernández-Lobato}

\ShortHeadings{Fixed-Mean Gaussian Processes for \emph{Post-hoc} Bayesian Deep Learning}{Ortega, Rodriguez-Santana and Hernández-Lobato}
\firstpageno{1}

\begin{document}

\title{Fixed-Mean Gaussian Processes for \emph{Post-hoc} Bayesian Deep Learning}
\author{\name Luis A. Ortega 
        \email laoa@cs.aau.dk
        \\ 
        \addr Universidad Autónoma de Madrid, Spain \& Aalborg University, Copenhagen.
        \AND 
        \name Simón Rodriguez-Santana 
        \email srsantana@icai.comillas.edu \\ 
        \addr Universidad Pontificia de Comillas, Spain
        \AND
        \name Daniel Hernández-Lobato
        \email daniel.hernandez@uam.es \\ 
        \addr Universidad Autónoma de Madrid, Spain.
        }

\editor{My editor}

\maketitle

\begin{abstract}
Recently, there has been an increasing interest in performing \emph{post-hoc} uncertainty estimation about the predictions of pre-trained deep neural networks (DNNs). 
Given a pre-trained DNN via back-propagation, these methods enhance the original network by adding output confidence measures, such as error bars, without compromising its initial accuracy. 
In this context, we introduce a novel family of sparse variational Gaussian processes (GPs), where the posterior mean is fixed to any continuous function when using a universal kernel. 
Specifically, we fix the mean of this GP to the output of the pre-trained DNN, allowing our approach to effectively fit the GP's predictive variances to estimate the 
DNN prediction uncertainty. Our approach leverages variational inference (VI) for efficient stochastic optimization, with training costs that remain independent of the number of training points, scaling efficiently to large datasets such as ImageNet. 
The proposed method, called fixed-mean GP (FMGP), is architecture-agnostic, relying solely on the pre-trained model's outputs to adjust the predictive variances. 
Experimental results demonstrate that FMGP improves both uncertainty estimation and computational efficiency when compared to state-of-the-art methods for DNN post-hoc Bayesian inference.
\end{abstract}


\section{Introduction}

Over the last years, deep neural networks (DNNs) have become the \textit{de-facto} solution for a range of pattern recognition problems due to their ability to model deterministic connections and obtain state-of-the-art generalization performance \citep{he2016deep}. However, DNNs suffer from significant disadvantages such as poorly calibrated probabilistic forecasts \citep{guo2017calibration} and poor reasoning ability in scenarios demanding model uncertainty \citep{blundell2015weight}. These issues are critical in risk-sensitive situations, e.g. autonomous driving \citep{kendall2017uncertainties} or healthcare \citep{leibig2017leveraging}.

Bayesian neural networks (BNNs) have successfully addressed the aforementioned issues in small-scale problems \citep{mackay1992practical, neal2012bayesian, graves2011practical}. However, employing these models in practical scenarios remains challenging, as they typically involve high-dimensional, multi-modal posterior distributions over the space of neural network parameters. Moreover, due to the intractability of the calculations required, the exact posterior in large BNNs is generally approximated through diverse inference techniques, including variational inference (VI) \citep{blundell2015weight}, Markov chain Monte Carlo (MCMC) \citep{chen2014stochastic} and the Laplace approximation (LA) \citep{mackay1992bayesian, ritter2018scalable}, among others. Nevertheless, empirical results often reveal a loss in predictive performance compared to simple DNNs trained via back-propagation \citep{wenzel20a}.

Recently, approaches based on the Linearized Laplace Approximation (LLA) \citep{immer2021improving}, 
which applies the Laplace Approximation to a linearized version of the DNN, have gained significant popularity. Their \emph{post-hoc} nature ensures that model performance is preserved. Specifically, LLA methods enhance the output of the DNN with the corresponding error bars that quantify prediction uncertainty. Notwithstanding, they demand computing the Jacobian of the DNN with respect to the parameters for each input of the training dataset, which is computationally expensive. Consequently, LLA methods often lack the scalability needed to be applied to large models and/or datasets.

In this work, we introduce a new family of sparse Gaussian processes (GPs), called \emph{fixed-mean Gaussian processes} (FMGPs). This approach leverages the dual representation of GPs in the Reproducing Kernel Hilbert Space (RKHS) and the concept of \emph{decoupled} inducing points for sparse GPs \citep{cheng2016incremental}. Specifically, by employing a universal kernel, our method enables fixing the posterior mean to any given continuous function. Then, it learns the corresponding posterior covariances of the model. As a result, the posterior mean can be set equal to the output of a high-performing DNN. VI is then used to stochastically optimize the GP’s predictive variances, providing useful error bars around the DNN’s predictions. The proposed method, FMGP, effectively \emph{converts any pre-trained DNN into a Bayesian DNN} through function-space inference. The two main advantages of this approach are: (i) it is scalable to large neural networks, as it avoids requiring DNN Jacobians and is less affected by the number of DNN parameters, and (ii) it leverages function-space inference for improved uncertainty estimation.

%Importantly, adapting the \emph{decoupled} GP framework of \cite{cheng2016incremental} for estimating only predictive variances and considering predictive means given by a DNN is not straight-forward. More precisely, in standard sparse GPs,  tuning hyperparameters involves balancing the fit of the mean to the data versus reducing the model's predictive uncertainty. However, in FMGP we fix the predictive mean, which eliminates this trade-off. Thus, the GP hyperparameters only adjust the predictive variance without affecting the mean. Consequently, optimizing them by maximizing the typical ELBO of VI leads to undesirable solutions where the predictive variance is set to zero. To avoid this and obtain good model hyperparameters, we introduce in FMGP a specific regularization technique.

The main benefit of using FMGP is its \emph{post-hoc} nature, where the pre-trained model predictions are preserved as the posterior mean of the GP, ensuring high performance. Additionally, compared to other \emph{post-hoc} approaches, the key advantage of FMGP is its architecture-agnostic design, as it relies solely on the DNN's outputs to accurately learn the predictive variances. This contrasts with methods such as LLA and its variants \citep{deng2022accelerated,ortega2024variational}, or mean-field approaches based on VI and fine-tuning \citep{deng2023bayesadapter}, which become computationally prohibitive for very large models due to requiring (i) high-dimensional DNN Jacobians or (ii) direct interaction with the parameters of the DNN. Further details on these methods and their differences with respect to FMGP are provided in Section~\ref{sec:related_work}.

\begin{figure}[t]
	\begin{center}
	\begin{tabular}{ccc}
    {\small LLA }& {\small FMGP }& {\small MFVI} \\
	\includegraphics[width=0.30\textwidth]{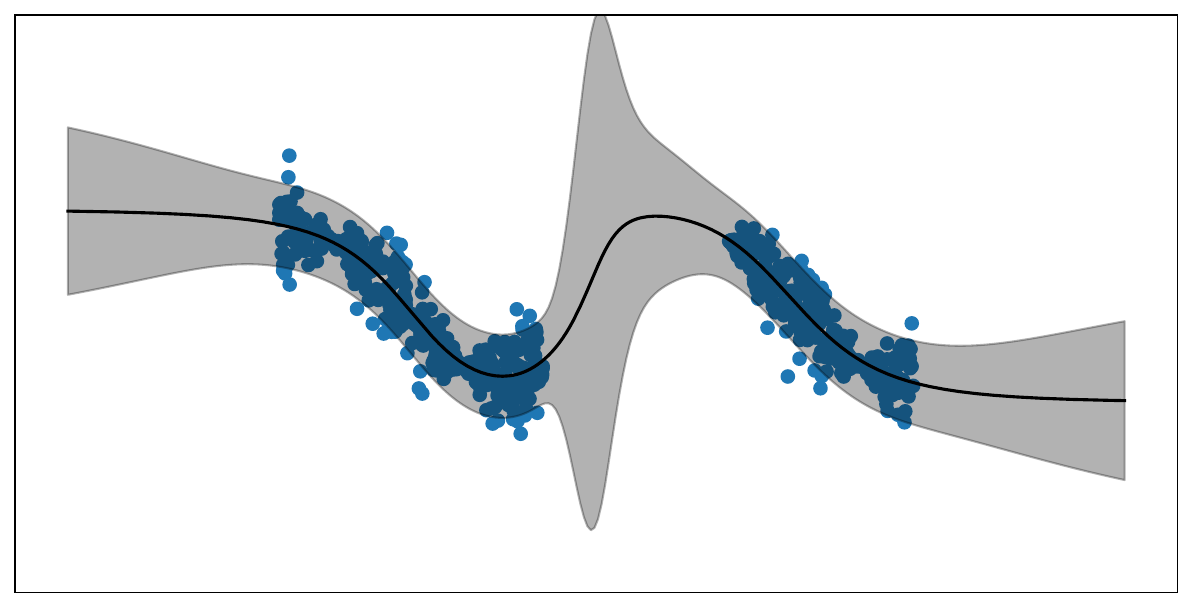} & \includegraphics[width=0.30\textwidth]{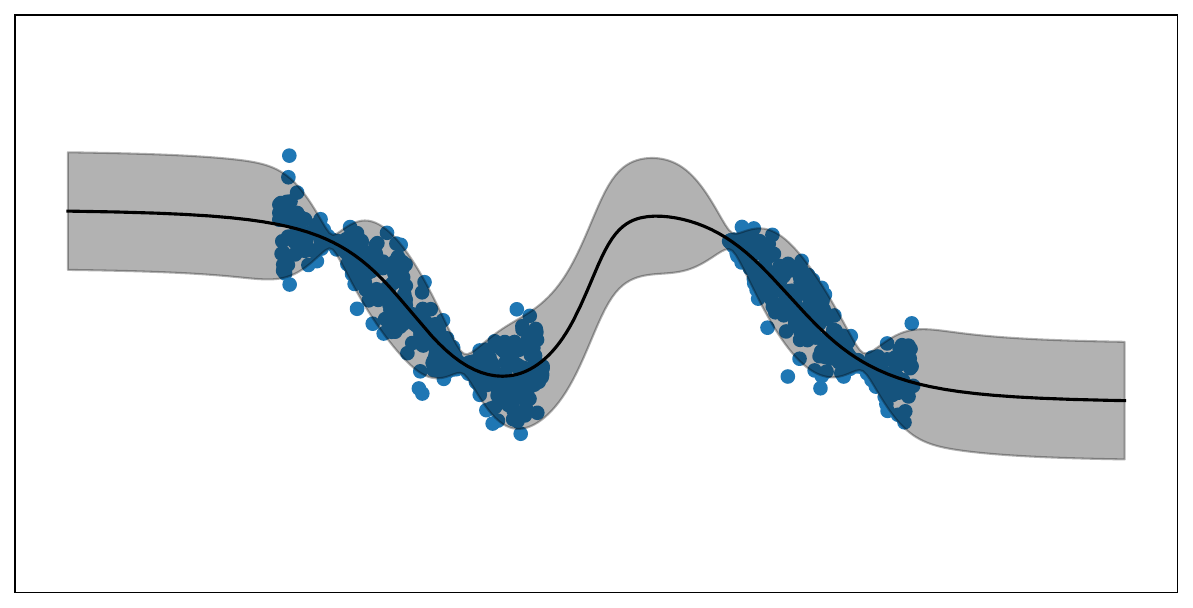} & \includegraphics[width=0.30\textwidth]{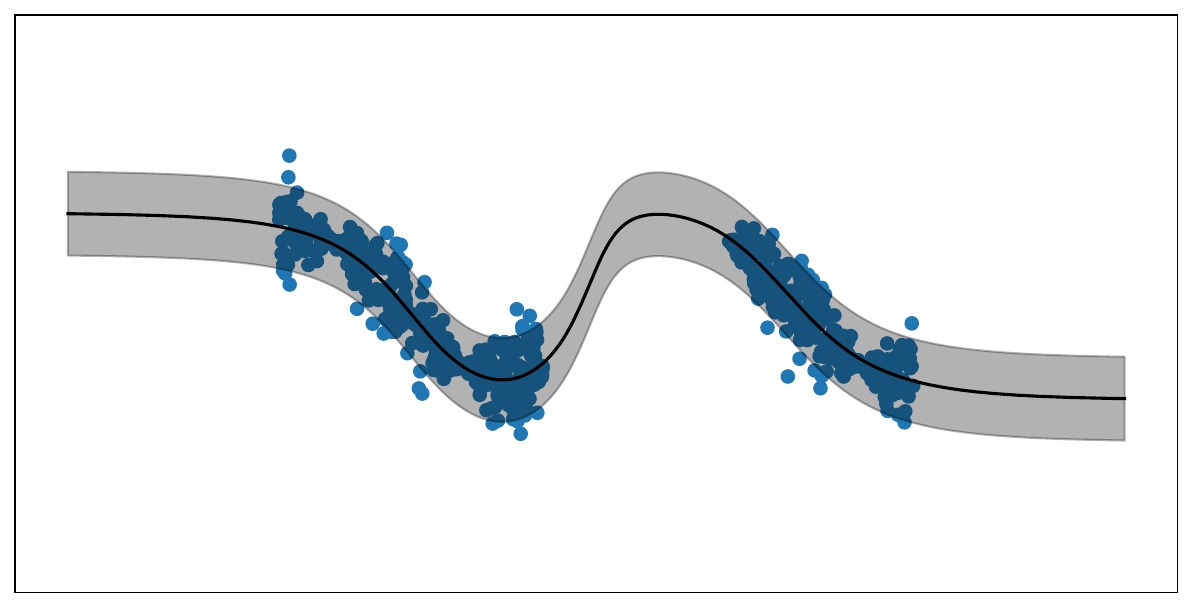} \\
    {\small GP Zero }& {\small GP } & {\small HMC}\\
	\includegraphics[width=0.30\textwidth]{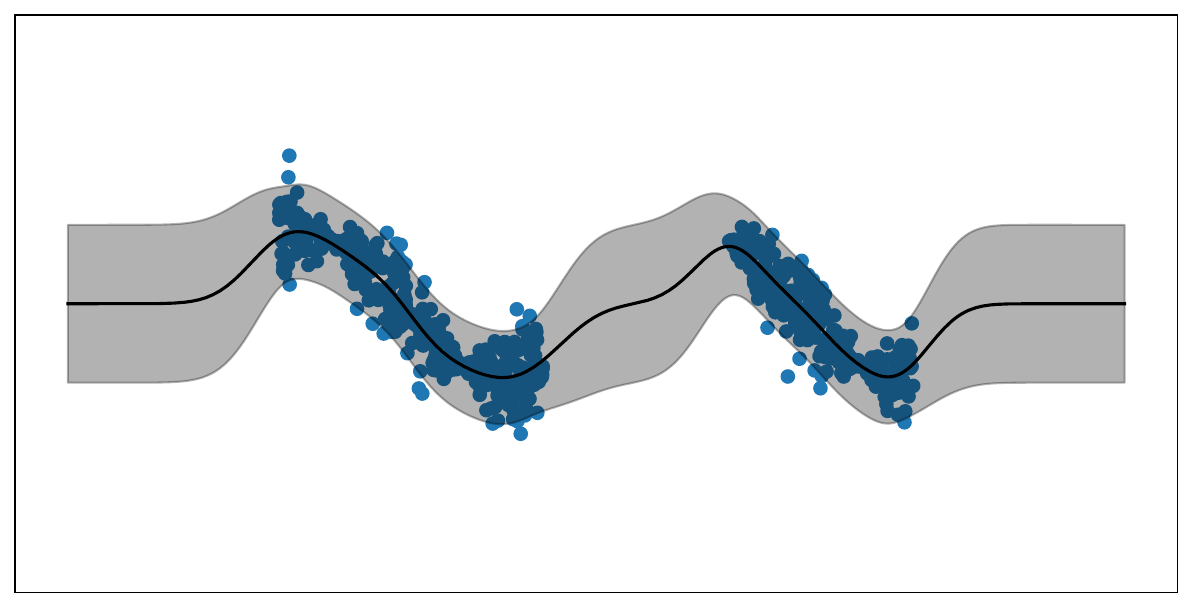} & \includegraphics[width=0.30\textwidth]{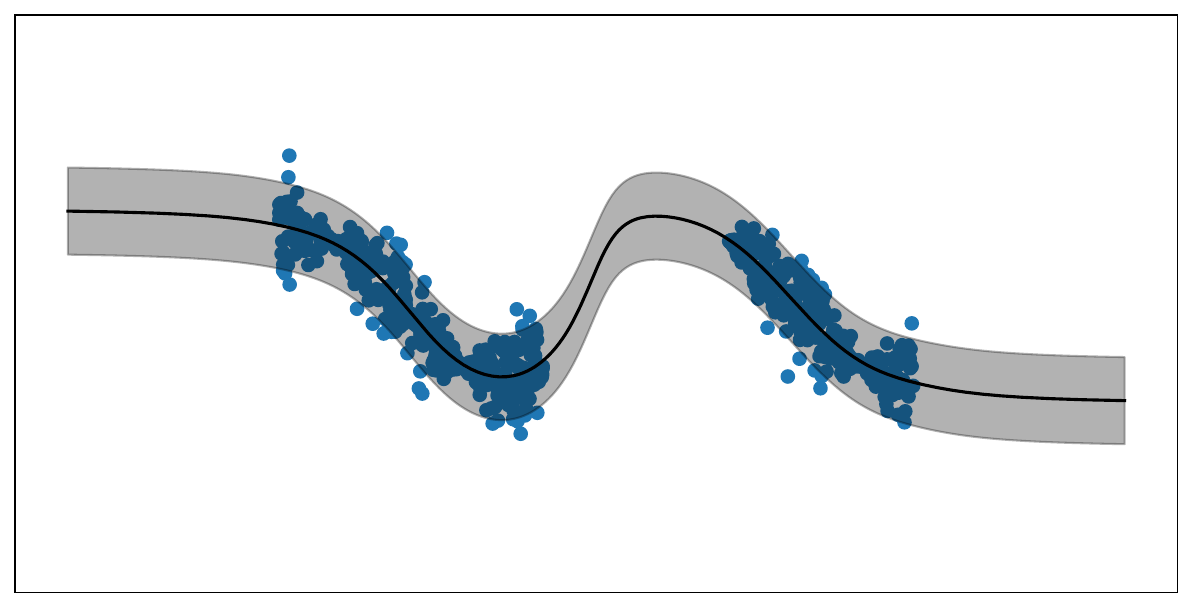} & \includegraphics[width=0.3\textwidth]{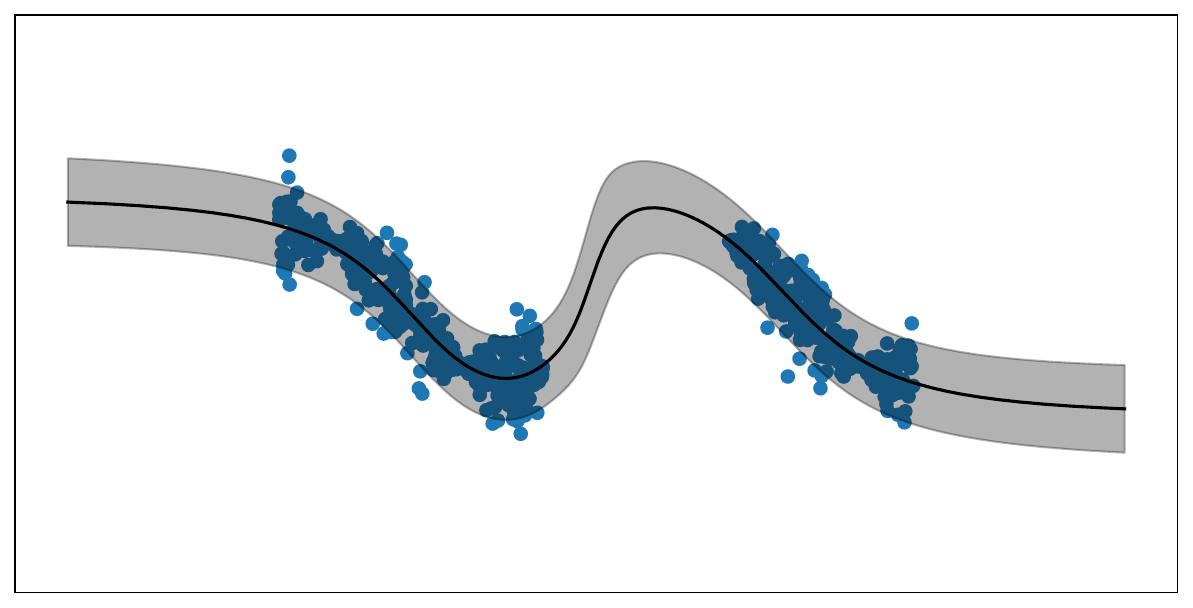}
	\end{tabular}
	\end{center}
    \caption{Predictive distribution (mean in black, \(2\sigma\) shaded region) on a toy 1D regression dataset. The backbone model is a \(2\) hidden layer MLP with \(50\) units trained using back-propagation. The compared approaches are linearized Laplace approximation (LLA), fixed-mean Gaussian process (FMGP) with squared exponential kernel, mean-field variational inference (MFVI) for DNN fine-tuning, Gaussian process (GP) with squared exponential kernel and zero mean prior (GP Zero); and the same GP with the MAP solution as GP prior mean (GP MAP), and Hamilton Monte Carlo (HMC). All methods' hyperparameters are optimized using training data except HMC, which uses uniform hyper-priors.}
    \label{fig:intro}
\end{figure}

Figure~\ref{fig:intro} illustrates the predictive distributions obtained by different methods on a toy 1-dimensional regression problem. We observe that the predictive distribution of FMGP avoids the uncertainty overestimation of LLA and is capable of modeling input-dependent noise. In contrast, the predictive distributions of other methods from the literature are unable to correctly model the sinusoidal nature of the input dependent noise, presenting high model bias. Further details about this experiment are provided in Section \ref{seq:synthetic_exp}.

\subsection*{Contributions}
The main contributions of this work are the following, we
\begin{enumerate}[label=\roman*), topsep=0pt, itemsep=-0.ex, partopsep=1ex, parsep=1ex]
    \item define FMGP as a family of GPs that can be used to perform uncertainty estimation in pre-trained DNNs without losing prediction performance;
    \item show how VI can be used to stochastically fit only the predictive variances;
    \item propose \(\alpha\)-divergences to efficiently optimize the model's hyperparameters while avoiding pathological solutions in which the predictive variances are set to zero. This results in a \emph{post-hoc} method that is independent of network structure or architecture (\emph{e.g.}, it does not require computing DNN Jacobians).
    \item show the scalability of FMGP at training and test time across multiple regression and classification problems, including ResNet \citep{he2016deep} models, with millions of parameters, and the ImageNet dataset, featuring thousands of class labels and millions of data instances \citep{ILSVRC15}. 
    \item show that FMGP can be used in the context of \emph{black-box classifiers} using CLIP networks \citep{radford2021learning}, without access to model parameters or training samples; 
    \item illustrate the utility of FMGP in a \emph{practical dataset} scenario using the QM9 dataset \citep{ruddigkeit2012enumeration} for feature prediction in the context of molecules.
\end{enumerate}

\section{Background}

In deep learning (DL), we aim to infer an unknown function \(h:\mathbb{R}^D \to \mathbb{R}\), based on noisy
observations \(\mathbf{y} = (y_1, \dots, y_N)^\text{T}\) at known locations \(\mathbf{X} = (\mathbf{x}_1, \dots, \mathbf{x}_N)\).
DL tackles this by choosing a neural network architecture, which induces a family of candidate functions
parameterized by learnable weights \(\bm\theta \in \mathbb{R}^P\),
\[
\mathcal{F} = \{f_{\bm{\theta}}:\mathbb{R}^D \to \mathbb{R}\}\,.
\]
The underlying assumption is that, if \(\mathcal{F}\) is large enough, some element in \(\mathcal{F}\) will closely approximate \(h\). That is,
\[
\exists f \in \mathcal{F}\quad \text{s.t.} \quad f_{\bm{\theta}^\star}(\cdot) \approx h(\cdot)\,.
\]
DL typically learns \(\bm{\theta}\) from data via back-propagation, yielding a point estimate \(\bm{\theta}^\star\) (and hence a single predictor \(f_{\bm{\theta}^\star}\)).
Nevertheless, despite its success in various tasks \citep{Vaswani2017}, standard DL methods generally lack proper output uncertainty estimation,
often resulting in over-confident predictions in regions without training data, where the uncertainty around \(g\) is expected to be larger.

In Bayesian inference, the observations \(\mathbf{y} = (y_1, \dots, y_N)^\text{T}\) are related to the target function evaluations \(\mathbf{f} = (f(\mathbf{x}_1), \dots, f(\mathbf{x}_N))^\text{T}\) through a likelihood function \(p(\mathbf{y} | \mathbf{f})\). In regression settings, where \(y_i \in \mathbb{R}\), the likelihood is often a homoscedastic Gaussian with variance $\sigma^2$. In classification, where \(y_i \in \{1, \dots, C\}\), the likelihood is categorical with class probabilities given by \emph{e.g.} a softmax activation function, meaning \(\mathcal{F} \subset \{f_{\bm{\theta}}:\mathbb{R}^D \to \mathbb{R}^C\}\) represents a set of multi-output functions, one per class label.

Bayesian neural networks (BNNs) follow a probabilistic framework \citep{mackay1992bayesian}, placing a prior over the network parameters \(p(\bm{\theta})\) and computing the Bayesian posterior 
\[
p(\bm{\theta} | \mathbf{y}) \propto p(\mathbf{y} | \bm{\theta})p(\bm{\theta})
\]
for predictions. Due to the non-linear nature of DNNs, calculating the posterior analytically is intractable. Therefore, most methods rely on an approximate posterior \(q(\bm{\theta}) \approx p(\bm{\theta} | \mathbf{y})\), used to estimate predictions via Monte Carlo sampling 
\[
p(y^\star | \mathbf{x}^\star, \mathbf{y}) = \mathbb{E}_{p(\bm{\theta} | \mathbf{y})}\left[p(y^\star | \mathbf{x}^\star, \bm{\theta})\right] \approx \mathbb{E}_{q(\bm{\theta})}\left[p(y^\star | \mathbf{x}^\star, \bm{\theta})\right] \approx S^{-1} \sum_{s=1}^S p(y^\star | \mathbf{x}^\star, \bm{\theta}_s)\,,
\]
where \(\bm{\theta}_s \sim q(\bm{\theta})\) and \(S\) represents the number of Monte Carlo samples. This allows effectively capturing the uncertainty in the model's predictions \citep{bishop2006}.

In this work, rather than following the Bayesian approach in the space of parameters, we take on a \emph{function-space} perspective. This involves placing a prior directly over the space of functions \(p(f)\) and constructing an approximate posterior for the target function \(q(f) \approx p(f | \mathbf{y})\) to make predictions 
\[
p(y^\star | \mathbf{x}^\star, \mathbf{y}) = \mathbb{E}_{p(f | \mathbf{y})}\left[p(y^\star | \mathbf{x}^\star, f)\right] \approx \mathbb{E}_{q(f)}\left[p(y^\star | \mathbf{x}^\star, f)\right]\,.
\]
These predictions can be computed exactly in regression settings using GP posteriors. In the case of classification problems, given an approximate GP posterior, they have to be approximated via Monte Carlo sampling.

\subsection{Gaussian Processes}

Gaussian processes (GPs) are statistically defined as an infinite collection of random variables such that any finite subset is jointly Gaussian. They are fully specified via mean and covariance functions. 

From this definition, GPs can be interpreted as distributions over the function space or, more precisely, over the set of evaluations of functions. Consider a function \(f:\mathbb{R}^D\to \mathbb{R}\). We say that \(f\) follows a GP defined by a mean function \(m(\cdot)\) and covariance (or kernel) function \(K(\cdot,\cdot)\), \emph{i.e.}, \(f \sim \mathcal{G}\mathcal{P}(m, K)\), if, for any finite set of input points \(\mathbf X = (\mathbf x_1,\dots,\mathbf x_N)^T \subset \mathbb{R}^D\), the set of function evaluations \(\mathbf f = (f(\mathbf x_1),\dots,f(\mathbf x_N))^T\) follows a multi-variate Gaussian distribution with mean $m(\mathbf{X})$ and covariance matrix $K(\mathbf{X}, \mathbf{X})$. That is, 
\[
f \sim \mathcal{G}\mathcal{P}(m, K) \iff \mathbf{f} = f(\mathbf{X}) \sim \mathcal{N}(m(\mathbf{X}), K(\mathbf{X}, \mathbf{X})), \quad \forall\, \mathbf{X} \subset \mathbb{R}\,.
\]

\subsection{Dual formulation of Gaussian Processes in RKHS}
\label{subsec:dual_formulation}

A Reproducing Kernel Hilbert Space (RKHS) \(\mathcal{H}\) is a Hilbert space of functions with the reproducing property, that is, 
\[
\forall \mathbf{x} \in \mathcal{X}, \ \exists \phi_{\mathbf{x}} \in \mathcal{H} \ \  \text{such that}\ \ \forall f \in \mathcal{H} \ \ \text{it verifies that} \ \ f(\mathbf{x}) = \braket{\phi_{\mathbf{x}}, f}_\mathcal{H}\,,
\]
where $\braket{\cdot,\cdot}_\mathcal{H}$ is the inner product on $\mathcal{H}$. By Moore–Aronszajn theorem \citep{aronszajn1950theory}, if \(K\) is a positive definite kernel on \(\mathcal{X}\), then, there exists a unique Hilbert space of functions on \(\mathcal{H}\) for which \(K\) is a reproducing kernel. More precisely, let \(\mathcal{H}_0(\mathcal{X})\) be the linear span of \(K\) on \(\mathcal{X}\) defined as
\begin{align}
	\mathcal{H}_0(\mathcal{X}) & := \left\{\sum_{i=1}^n a_i \phi_{\mathbf{x}_i}:\, n \in \mathbb{N}, \ a_i \in \mathbb{R}, \ \mathbf{x}_i \in \mathcal{X}\right\}\,.
\end{align}
By Moore–Aronszajn theorem, the closure of \(\mathcal{H}_0(\mathcal{X})\), named as \(\mathcal{H} := \overline{\mathcal{H}_0(\mathcal{X})}\) is a Hilbert space verifying the reproducing property with \(\phi_\mathbf{x} = K(\cdot, \mathbf{x}), \ \forall \mathbf{x} \in \mathcal{X}\).

A \(\mathcal{GP}(m, K)\) has a dual representation as a Gaussian measure in a Banach space that contains the RKHS of its kernel function \citep{holmes2015gaussian, cheng2016incremental}. More precisely, consider a zero-mean GP prior with \(m=0\) and the RKHS defined by the kernel \(K\) associated to the GP. For any \(\mu \in \mathcal{H}\) and a linear semi-definite positive 
operator \(\Sigma\) associated to $\mathcal{H}$, \emph{i.e}, 
\(\Sigma \in \mathcal{L}^+(\mathcal{H},\mathcal{H})\), we can define a new GP with mean function \(m^\times\) and kernel function \(K^\times\) given by
\begin{equation}\label{eq:gp_dual}
	m^\times(\mathbf{x}) = \braket{\phi_{\mathbf{x}},\mu}_{\mathcal{H}}\,, \ \ \text{ and}, \ \ K^\times(\mathbf{x}, \mathbf{x}' ) = 
	\braket{\phi_{\mathbf{x}},\Sigma (\phi_{\mathbf{x}'})}_{\mathcal{H}}\,.
\end{equation}
We use \(p(f) = \mathcal{N}(f|\mu, \Sigma)\) as an abuse of notation to denote such Gaussian measure in the Banach space, with \(\mathcal{P}_{\mathcal{H}}\) as the set of these measures:
\begin{align}
	\mathcal{P}_{\mathcal{H}} & = \Big\{\mathcal{N}(f|\mu, \Sigma) \ : \ \mu \in \mathcal{H}, \ \Sigma \in \mathcal{L}^{+}(\mathcal{H}, \mathcal{H})\Big\}\,.
\end{align}
As a result, there is a correspondence between GPs \(\mathcal{GP}(m^\times, K^\times)\) and Gaussian measures \(\mathcal{N}(f|\mu, \Sigma)\) in a Banach space \(\mathcal{B}\) that contains the samples of the GP and in which  \(\mathcal{H}\) is dense \citep{holmes2015gaussian, cheng2017variational}.

\begin{remark}
    The zero-mean GP prior \(\mathcal{GP}(0, K)\) is obtained from the dual-formulation using \(\mathcal{N}(f|0, I)\). Furthermore, given a set 
    of observations \(\mathbf{y}\) from a regression task with Gaussian noise \(\sigma^2\), 
    the GP posterior is obtained from the (posterior) Gaussian measure 
    \(p(f|\mathbf{y}) = \mathcal{N}(f|\mu^\star, \Sigma^\star)\) 
    where:
    \begin{equation}
    	\mu^\star = \textstyle \sum_{i=1}^N \alpha_i \phi_{\mathbf{x}_i}\,, \quad
    	\Sigma^\star(\phi) = \textstyle \phi - \sum_{i=1}^N \sum_{j=1}^N \phi_{\mathbf{x}_i} \Lambda_{i,j} \braket{\phi_{\mathbf{x}_j}, \phi}_\mathcal{H}\,,
    \end{equation}
    with \(\bm{\Lambda} = (K(\mathbf{X}, \mathbf{X}) + \sigma^2 \bm{I})^{-1} \in \mathbb{R}^{N \times N}\) and \(\bm{\alpha} = \bm{\Lambda}\bm{y} \in \mathbb{R}^N\).
\end{remark}
With this construction, we aim to define a family of Gaussian measures in~\(\mathcal{P}_\mathcal{H}\), whose corresponding GP verifies that \(m(\mathbf{x}) \approx g(\mathbf{x})\). This means that the corresponding GP mean will match the output of the pre-trained neural network \(g(\cdot)\). Then, VI can be used to find an optimal Gaussian measure within such family.

\subsection{Universal Kernels}\label{app:universal}
Following \cite{micchelli2006universal}, we introduce the notion of \emph{universal kernels} as kernel functions whose linear span can approximate any continuous function in a compact set. Given a kernel function \(K(\cdot, \cdot)\) and its corresponding RKHS \(\mathcal{H}\), assume that the kernel is continuous on \(\mathcal{X} \times \mathcal{X}\). Let \(\mathcal{Z}\) be a fixed but arbitrary compact subset of \(\mathcal{X}\) and, as usual, let \(C(\mathcal{Z})\) denote the space of all continuous real-valued functions from \(\mathcal{Z}\) to \(\mathbb{R}\) equipped with infinity  norm \(\|\cdot\|_{\infty}\), which reduces to a \emph{maximum norm} in the compact set \(\|\cdot\|_{\mathcal{Z}}\).

The space of \emph{kernel sections} is defined as  \(K(\mathcal{Z}) := \overline{\mathcal{H}_0(\mathcal{Z})}\), which consists of the set of all continuous functions \(C(\mathcal{Z})\) which are limits of linear combinations of  \(\{K(\cdot, \mathbf z) \ : \mathbf{z} \in \mathcal{Z}\}\) under the infinity norm.

\begin{definition}\label{def:universal}
    A kernel function is said to be universal if for any compact subset \(\mathcal{Z}\) of the input space \(\mathcal{X}\), the kernel section \(K(\mathcal{Z})\) is dense in \(C(\mathcal{Z})\) with the infinity norm. That is, for any \(f \in C(\mathcal{Z})\) and any \(\epsilon > 0\), there exists \(g_\epsilon \in K(\mathcal{Z})\) such that \(\|f - g_\epsilon \|_\infty \leq \epsilon\).
\end{definition}
From the above definition, if \(K\) is universal, \( \forall f \in C(\mathcal{Z})\) and \(\epsilon > 0\), there exists a set of \(M \in \mathbb{N}\) scalar values \(a_1, \dots, a_M \in \mathbb{R}\) and input space points \(\{\mathbf{z}_1, \dots, \mathbf{z}_M\} \subset \mathcal{Z}\), such that
\begin{align}
    \Big\|f(\cdot) - \sum_{m=1}^{M}a_m K(\cdot, \mathbf{z}_m)\Big\|_{\infty} \leq \epsilon\,.
\end{align}
Intuitively, a universal kernel can approximate any continuous function in a compact set via linear combinations of kernel evaluations. As the approximation improves (\emph{i.e.} \(\epsilon\) decreases), the number of terms \(M\) needed in the linear combination increases.

The squared-exponential kernel with hyperparameters \(\Omega=\{\alpha, \{l_j\}_{j=1}^D\}\), with \(l_j \in \mathbb{R}^+\), defined as
\begin{align}
    K_\text{RBF}(\mathbf{x}, \mathbf{x}') & := \alpha \exp \left(-\frac{1}{2} \sum_{j=1}^D \frac{(x_j - x'_j)^2}{l_j}\right)\,,
\end{align}
is a universal kernel~\citep{micchelli2006universal}. 

\section{Fixed-Mean Gaussian Processes}

Here, we present a novel family of GPs, \textit{Fixed-Mean Gaussian Processes} (FMGPs). This family of function-space distributions is defined using the dual formulation of sparse variational GPs,
which are introduced next.

\subsection{Sparse Variational Gaussian Processes}

Sparse Variational GPs (SVGPs) \citep{titsias2009variational} approximate the GP posterior using a GP parameterized by $M$ inducing points 
$\mathbf{Z}=(\mathbf{z}_1,\ldots,\mathbf{z}_M)$, with each $\mathbf{z}_i \in \mathbb{R}^D$, and associated process values 
\(\mathbf{u}=(u_1,\ldots,u_M)^\text{T}:= f(\mathbf{Z})\). Specifically,
\begin{align*}
	p(\mathbf{f}, \mathbf{u} | \mathbf{y}) \approx q(\mathbf{f}, \mathbf{u}) & = p(\mathbf{f} | \mathbf{u})q(\mathbf{u})\,,
\end{align*}
where  \(q(\mathbf{u}) = \mathcal{N}(\mathbf{u}|\hat{\bm{m}}, \hat{\bm{S}})\), $\mathbf{f}=f(\mathbf{X})$ and $p(\mathbf{f} | \mathbf{u})$ is fixed
to the GP predictive distribution. 

Following \cite{cheng2016incremental}, consider a restriction of the dual GP formulation introduced in Section~\ref{subsec:dual_formulation},
where the mean and covariance dual elements (\(\mu\) and \(\Sigma\)) must satisfy the linear structure:
\begin{align*}
    \tilde{\mu}_{\bm a} & = \textstyle \sum_{m=1}^{M} a_m \phi_{\mathbf{z}_m}, \\
    \tilde{\Sigma}_{\bm{A}}(\phi) & = \textstyle \phi + \sum_{i=1}^{M}\sum_{j=1}^M \phi_{\mathbf{z}_i} A_{i,j} \braket{\phi_{\mathbf{z}_j}, \phi}_\mathcal{H}, 
\end{align*}
where \(\bm{a} = (a_1, \dots, a_M)^T \in \mathbb{R}^M\), \(\bm{A} = (A_{ij}) \in \mathbb{R}^{M \times M} \) such that \(\tilde{\Sigma} \geq 0\) and \(\phi_{\mathbf{z}} := K(\cdot, \mathbf{z}_i) \in \mathcal{H},\ \forall \mathbf{z} \in \mathbf{Z}\). This defines a family of Gaussian measures \(\mathcal{Q}\subset \mathcal{P}_{\mathcal{H}}\) such that
{
\begin{align*}
	\mathcal{Q} & = \left\{\mathcal{N}(f|\tilde{\mu}_{\bm a}, \tilde{\Sigma}_{\bm A}): \bm{a} \in \mathbb{R}^M, \bm{A} \in \mathbb{R}^{M \times M}, \mathbf Z \in \mathcal{X}^M \right\}\,,
\end{align*}
}where we have omitted \(\mathcal{H}\) and \(M\) from \(\mathcal{Q}'s\) notation for simplicity as more sub-indexes will be used later. 

\begin{proposition}
Let $\mathbf Z=(\mathbf z_1,\dots,\mathbf z_M)$ be inducing inputs, $\bm u=f(\mathbf Z)$, and
\[
q(\mathbf u)=\mathcal N(\mathbf u| \bm \mu,\bm S),\qquad p(\mathbf u)=\mathcal N(0,K(\mathbf Z, \mathbf Z))\,.
\]
Then the induced marginal $q(f)=\int p(f| \mathbf u)\,q(\mathbf u)\,\mathrm{d}\mathbf u$
belongs to the dual family $\mathcal Q$ (as defined above) and can be written
as $q(f)=\mathcal N\big(f|\tilde{ \mu}_{\bm{a}},\tilde{ \Sigma}_{\bm{A}}\big)$ with parameters
\[
\bm a = K(\mathbf Z, \mathbf Z)^{-1}\bm \mu,
\qquad
\bm A = K(\mathbf Z, \mathbf Z)^{-1} \bm S K(\mathbf Z, \mathbf Z)^{-1} - K(\mathbf Z, \mathbf Z)^{-1}.
\]
\end{proposition}
\begin{proof}
    For full derivations and related equivalences between inducing-point Gaussians and RKHS/dual representations, see~\citet{cheng2016incremental}.
    Under the GP prior, the conditional $p(f|\mathbf u)$ is Gaussian with mean linear in $\mathbf u$ and covariance equal to the prior covariance minus the Nystr\"om term induced by $\mathbf Z$. Since $q(\mathbf u)$ is Gaussian, marginalizing $u$ gives a GP $q(f)$ whose mean is a kernel expansion on $\mathbf Z$ with coefficients $K(\mathbf Z, \mathbf Z)^{-1}\bm{\mu}$. The covariance equals the prior covariance plus a rank-$M$ correction that can be written as $K(\mathbf x, \mathbf Z)\bm A K(\mathbf Z, \mathbf x')$, where matching terms yields $\bm A=K(\mathbf Z, \mathbf Z)^{-1}\bm S K(\mathbf Z, \mathbf Z)^{-1}-K(\mathbf Z, \mathbf Z)^{-1}$. Hence $q(f)$ lies in the dual family $\mathcal Q$ with parameters $(\bm a,\bm A)$ as stated. 
\end{proof}
By definition, if \(M = N\), then \(p(f|\mathbf{y}) \in \mathcal{Q}\), as in standard variational sparse GPs. However, in practice, and for scalability reasons, \(M \ll N\) and \(p(f|\mathbf{y}) \notin \mathcal{Q}\). This leads to the problem of finding the measure in \(\mathcal{Q}\) that is \emph{closest} to \(p(f|\mathbf{y})\). In this regard, VI can be used to minimize the KL divergence between Gaussian measures \citep{cheng2016incremental} and hence, to compute the \emph{optimal} variational measure as:
\begin{equation}
    \argmin_{q \, \in\,  \mathcal{Q}} \ \mathrm{KL}\big(q(f) \big| p(f|\mathbf{y})\big)
    = \argmax_{q\,  \in \, \mathcal{Q}} \ \mathbb{E}_{q(f)}[\log p(\mathbf{y}|f)]
    - \mathrm{KL}\big(q(f) \big| p(f)\big)\,,
    \label{eq:vi_dual_formulation}
\end{equation}
where \(\mathrm{KL}\big(q(f) \big| p(f)\big)\) can be computed in closed form 
with \(p(f)=\mathcal{N}(f|0, I)\), \emph{i.e.}, the GP prior. Namely, 
\begin{equation*}
        \mathrm{KL}\big(q(f) \big| p(f)\big) = 
	 \frac{1}{2} \bm{a}^T K(\mathbf{Z}, \mathbf{Z}) \bm{a}  + \frac{1}{2} \text{tr}\left( K(\mathbf{Z}, \mathbf{Z})\bm{A} \right)+ \frac{1}{2} \log |\bm{M}| \,,
\end{equation*}
with $\bm{M}=\bm{I} - K(\mathbf{Z}, \mathbf{Z})(\bm{A}^{-1} + K(\mathbf{Z}, \mathbf{Z}))^{-1}$. After optimizing Equation~\eqref{eq:vi_dual_formulation}, one gets a Gaussian measure $q$ that corresponds to the SVGP in \cite{titsias2009variational}. See \cite{cheng2017variational} for further details about this.

\subsection{Decoupled Basis}

In \cite{cheng2017variational}, the authors propose to generalize \(\mathcal{Q}\) so that \(\tilde{\mu}_{\bm a}\) and \(\tilde{\Sigma}_{\bm A}\) are defined using different sets of inducing points. Let \(\mathbf{Z}_{\alpha} \in \mathcal{X}^{M_\alpha}\) and \(\mathbf{Z}_{\beta} \in \mathcal{X}^{M_\beta}\) be two sets of inducing points, for the mean and the variance respectively, of sizes $M_\alpha$ and $M_\beta$. The generalized dual representation is then defined as:
\begin{align*}
    \tilde{\mu}_{\alpha, \bm a} &= \textstyle \sum_{m=1}^{M_\alpha} a_m \phi_{\mathbf{z}_{\alpha, m}}\,,\\
	\tilde{\Sigma}_{\beta, \bm{A}}(\phi) &= \textstyle \phi + \sum_{i=1}^{M_ \beta}\sum_{j=1}^{M_\beta}\phi_{\mathbf{z}_{\beta, i}} A_{i,j} \braket{\phi_{\mathbf{z}_{\beta, j}}, \phi}_\mathcal{H}\,.
\end{align*}
This decoupled parameterization is a clear generalization from standard SVGPs and cannot be obtained using the approach of \cite{titsias2009variational} unless \(\mathbf{Z}_\alpha = \mathbf{Z}_\beta\). 
The decoupled space of Gaussian measures is now:
{
\begin{align*}
\mathcal{Q}^{+} = \left\{\mathcal{N}(f|\tilde{\mu}_{\alpha, \bm a}, \tilde{\Sigma}_{\beta, \bm A}) : \bm{a} \in \mathbb{R}^{M_\alpha},\bm{A} \in \mathbb{R}^{M_\beta \times M_\beta}, \mathbf Z_\alpha  \in \mathcal{X}^{M_\alpha}, \mathbf Z_\beta  \in \mathcal{X}^{M_\beta}\right\}\,,
\end{align*}
}where it verifies that \(\mathcal{Q} \subset \mathcal{Q}^{+}\). As shown in~\cite{cheng2017variational}, VI can be 
used to find the optimal $q$ in this parametric family. That is,
\begin{align}
    \argmin_{q\, \in\, \mathcal{Q}^+} \ \mathrm{KL}\big(q(f) \big| p(f|\mathbf{y})\big)  = \argmax_{q\, \in\, \mathcal{Q}^+} \ \mathbb{E}_{q(f)}[\log p(\mathbf{y}|f)] - \mathrm{KL}\big(q(f) \big| p(f)\big)
	\label{eq:vi_decoupled}
	\,,
\end{align}
where the KL term, with \(p(f)=\mathcal{N}(f|0, I)\), is:
\begin{align}
        \mathrm{KL}\big(q(f) \big| p(f)\big) &= \frac{1}{2} \bm{a}^T \bm{K}_{\alpha} \bm{a}  + \frac{1}{2} \text{tr}\left( \bm{K}_{\beta}\bm{A} \right) + \frac{1}{2} \log |\bm{I} - \bm{K}_{\beta}(\bm{A}^{-1} + \bm{K}_{\beta})^{-1}|\,,
	\label{eq:opt_kl}
\end{align}
with \(\bm{K}_{\alpha} = K(\mathbf{Z}_\alpha, \mathbf{Z}_\alpha)\) and \(\bm{K}_\beta = K(\mathbf{Z}_\beta, \mathbf{Z}_\beta)\).\\

\begin{remark}
	The parameters for the mean of the variational distribution, \emph{i.e.}, \(\mathbf{Z}_\alpha\) and \(\bm a\), and the ones for the variance, \emph{i.e.}, \(\mathbf{Z}_\beta\) and \(\bm A\), are separated in Equations~\eqref{eq:vi_decoupled} and \eqref{eq:opt_kl}. Therefore, they can be independently optimized in practice.
\end{remark}

\subsection{Fixed-Mean Variational Family}

Given a universal kernel, we now use the decoupled family of distributions to show that the mean function can be \emph{fixed} to any continuous function in compact subsets. We recall the definition and key properties of universal kernels in Section~\ref{app:universal}.

\begin{proposition}\label{prop:fixed_mean}
    Let \(\mathcal{Z} \subset \mathcal{X}\) be any compact subset of the input space. If the kernel is \emph{universal}, for any function \(g \in C(\mathcal{Z})\), there exists \(M_\alpha > 0\), a set of inducing locations \(\{\mathbf{z}_1, \dots, \mathbf{z}_{M_\alpha}\} \subset \mathcal{Z}\), and scalar values \(a_1, \dots, a_{M_\alpha} \in \mathbb{R}\) such that
	\begin{align*}
		m(\mathbf{x}) & = \braket{\phi_{\mathbf{x}}, \tilde{\mu}_{\alpha, \bm{a}}} = \sum_{m=1}^{M_\alpha}a_m K(\mathbf{x}, \mathbf{z}_m) \,,
	\end{align*}
    verifies $\big\|g(\mathbf{x}) -  m(\mathbf{x})\big\|_{\mathcal{Z}} \leq \epsilon$.
\end{proposition}
\begin{proof}
For a compact set 
\(\mathcal{Z} \subset \mathcal{X}\), define the kernel section
\[
    K(\mathcal{Z}) := \Big\{ 
        \sum_{m=1}^M a_m K(\cdot,\mathbf{z}_m) 
        : M \in \mathbb{N},\, a_m \in \mathbb{R},\, \mathbf{z}_m \in \mathcal{Z}
    \Big\} \subset C(\mathcal{Z}).
\]
By universality, \(K(\mathcal{Z})\) is dense in \(C(\mathcal{Z})\) with respect to the 
sup–norm. Thus, for the given \(g \in C(\mathcal{Z})\) and \(\epsilon > 0\), there exists
a function \(m \in K(\mathcal{Z})\) such that
\[
    \|g - m\|_{\mathcal{Z}} \le \epsilon.
\]
By the definition of \(K(\mathcal{Z})\), this \(m\) can be written as a finite linear
combination
\[
    m(\mathbf{x}) = \sum_{m=1}^{M_\alpha} a_m K(\mathbf{x}, \mathbf{z}_m),
    \qquad \mathbf{z}_m \in \mathcal{Z},\ a_m \in \mathbb{R}.
\]
Let \(\mathcal{H}\) be the RKHS associated with \(K\) and \(\phi: \mathcal{X} \to \mathcal{H}\)
a feature map such that \(K(\mathbf{x},\mathbf{z}) 
= \langle \phi_{\mathbf{x}}, \phi_{\mathbf{z}} \rangle_{\mathcal{H}}\). Define
\[
    \tilde{\mu}_{\alpha,\bm{a}} := \sum_{m=1}^{M_\alpha} a_m \phi_{\mathbf{z}_m} \in \mathcal{H}.
\]
Then, for any \(\mathbf{x} \in \mathcal{Z}\),
\[
    \braket{\phi_{\mathbf{x}}, \tilde{\mu}_{\alpha,\bm{a}}}
    = \sum_{m=1}^{M_\alpha} a_m 
      \braket{\phi_{\mathbf{x}}, \phi_{\mathbf{z}_m}}_{\mathcal{H}}
    = \sum_{m=1}^{M_\alpha} a_m K(\mathbf{x}, \mathbf{z}_m)
    = m(\mathbf{x}).
\]
Hence \(m(\mathbf{x}) = \braket{\phi_{\mathbf{x}}, \tilde{\mu}_{\alpha,\bm{a}}}\) has the 
desired form and satisfies \(\|g - m\|_{\mathcal{Z}} \le \epsilon\). Since \(g\) and \(\epsilon\) were arbitrary, the claim follows.
\end{proof}

\noindent As a result, given an error rate \(\epsilon > 0\), we can set the posterior mean of a decoupled GP to \emph{any continuous function in any compact set of the input space}. More precisely, we can use the decoupled formulation of GPs to \textit{fix} the posterior mean to the output $g(\cdot)$ of a given pre-trained DNN.

\begin{definition}\label{def:q}
    For any compact subset of the input space \(\mathcal{Z} \subset \mathcal{X}\), continuous function \(g \in C(\mathcal{Z})\), error \(\epsilon > 0\) and universal kernel \(K \in C(\mathcal{X} \times \mathcal{X})\), the set of \(g\)-mean Gaussian measures $\mathcal{Q}^{g}_{\mathcal{Z}, \epsilon} \subset \mathcal{Q}^+$ is defined as
    \begin{align*}
	    \mathcal{Q}^{g}_{\mathcal{Z}, \epsilon} & := \left\{\mathcal{N}(f| \tilde{\mu}_{\alpha, \bm{a}}, \tilde{\Sigma}_{\beta, \bm A}) \ : \
            \bm{A} \in \mathbb{R}^{M_\beta \times M_\beta},
            \mathbf Z_\beta \in \mathcal{X}^{M_\beta}\right\} \,,
    \end{align*}
	where \(\tilde{\mu}_{\alpha, \bm{a}}\) verifies \(\big\|g(\mathbf{x}) -  \braket{\phi_{\mathbf{x}}, \tilde{\mu}_{\alpha, \bm{a}}}_\mathcal{H}\big\|_{\mathcal{Z}} \leq \epsilon\).
    \end{definition}
    \noindent Thus, for any \(q(f) \in  \mathcal{Q}^{g}_{\mathcal{Z}, \epsilon}\), the corresponding GP \(f \sim \mathcal{GP}(m^{\star}, K^\star_{\mathbf{Z}_\beta, \bm{A}})\) verifies that
    \begin{equation*}
	    \big\|g(\mathbf x) - m^\star(\mathbf{x}) \big\|_{\mathcal{Z}} \leq \epsilon\,,
    \end{equation*}
    and
    \begin{equation}
    \label{eq:pred}
	    K^{\star}_{\mathbf{Z}_\beta, \bm{A}}(\mathbf{x}, \mathbf{x}')  = K(\mathbf{x}, \mathbf{x}') +  K(\mathbf{x}, \mathbf{Z}_\beta)\bm{A}K(\mathbf{Z}_\beta, \mathbf{x}') \,.
    \end{equation}
    We will refer to this set of GPs as fixed-mean Gaussian processes (FMGPs). By Proposition~\ref{prop:fixed_mean}, it is clear that for any \(g \in C(\mathcal{Z})\), it is verified that \(\mathcal{Q}^{g}_{\mathcal{Z}, \epsilon} \neq \emptyset\) and its 
    corresponding set of FMGPs exists. Again, VI can be used to find the optimal $q$ from this parametric family.
	That is,
	\begin{align}
	    \argmin_{q\,  \in\, \mathcal{Q}^{g}_{\mathcal{Z}, \epsilon}} \ \mathrm{KL}\big(q(f) \big| p(f|\mathbf{y})\big)  &= \argmax_{q \,\in \,\mathcal{Q}^g_{\mathcal{Z}, \epsilon}} \ \mathbb{E}_{q(f)}[\log p(\mathbf{y}|f)] - \mathrm{KL}\big(q(f) \big| p(f)\big)\,,
 \label{eq:opt_kl2}
	\end{align}
    where the KL term is, setting \(p(f)=\mathcal{N}(f|0, I)\): 
\begin{align*}
        \mathrm{KL}\big(q(f) | p(f)\big) &= \frac{1}{2}  \log |\bm{I} - \bm{K}_\beta(\bm{A}^{-1} + \bm{K}_\beta)^{-1}|- \frac{1}{2} \text{tr}\left( \bm{K}_\beta\bm{A} \right) + \text{constant}\,.
\end{align*}
Note now, however, that only $\mathbf{A}$ and $\mathbf{Z}_\beta$ need to be optimized.

\begin{remark}
    Parameterizing \(\tilde{\Sigma}_{\bm{A}}\) with \(\bm{A} = -(\tilde{\bm{A}}^{-1} + \bm{K}_\beta)^{-1}\) results in 
    an expression for the posterior covariances in Equation~\eqref{eq:pred} equivalent to the expression for the posterior 
	covariances in the SVGP \citep{titsias2009variational}. Furthermore, Equation~\eqref{eq:pred} verifies that the posterior 
	covariances are less confident than the prior covariances \citep{cheng2017variational}.
\end{remark}

% \noindent Figure~\ref{fig:diagram} shows a set representation of the different families of Gaussian measures
% considered in this section. We observe that $\mathcal{Q}^+$, in which there are different inducing
% points for the mean and the covariances, is the largest family. This family includes
% both $\mathcal{Q}$, in which there is only a single set of inducing points for the mean and the covariances,  
% and $\mathcal{Q}^{g}_{\mathcal{Z}, \epsilon}$, in which the posterior mean is fixed to approximate 
% $g$ in $\mathcal{Z}$ with error at most $\epsilon$. Note that there could potentially 
% be some overlap between the Gaussian measures in $\mathcal{Q}$ and in $\mathcal{Q}^{g}_{\mathcal{Z}, \epsilon}$.

% \begin{figure}[t]
% 	\begin{center}
% 	\includegraphics[width=0.4\textwidth]{imgs/diagram2.pdf}
% 	\end{center}
%     \caption{Representation of the considered sets of variational Gaussian measures for fixed-mean Gaussian processes.}
%     \label{fig:diagram}
% \end{figure}

\subsection{Application to \emph{Post-hoc} Bayesian Deep Learning}

FMGP enables the conversion of DNNs into approximate Bayesian models while 
maintaining the DNN output as the predictive mean. The process can 
be summarized as follows:
\begin{enumerate}[itemsep=-0.ex, partopsep=1ex, parsep=1ex]
    \item Given a pre-trained model \(g \in C(\mathcal{X})\), choose a parametric family of kernels that defines an RKHS and a family of Gaussian measures \(\mathcal{P}_\mathcal{H}\), \emph{e.g.} squared exponential.
    \item Ensure that there exists a compact set \(\mathcal{Z} \subset \mathcal{X}\) where inputs are expected. The parametric family of Gaussian measures \(\mathcal{Q}^{g}_{\mathcal{Z}, \epsilon}\) exists for any \(\epsilon > 0 \).
    \item Initialize a measure in this family, \emph{e.g.} initialize \(\mathbf{Z}_\beta\) using K-Means \citep{lloyd1982least} and \(\tilde{\bm{A}}\) as the identity matrix.
    \item Perform VI to optimize the variational measure (\(\bm{A}\) and \( \mathbf{Z}_\beta\)), along with the kernel hyperparameters $\Omega$ and the noise variance $\sigma^2$, using (\ref{eq:opt_kl2}). The predictive variance is computed as in \eqref{eq:pred}, and, if \(\mathcal{Z}\) is large and \(\epsilon\) small, the predictive mean \(m^\star\) approximates the pre-trained model \(g\). Thus, in practice, $m^\star$ can be replaced by $g$ in the computations.
\end{enumerate}

\subsection{Loss Function}

In standard sparse GPs, tuning hyperparameters involves balancing the fit of the mean 
to the training data versus reducing the model's predictive variance. However, FMGPs 
fix the predictive mean, which eliminates this trade-off. Thus, the kernel hyperparameters \(\Omega\) 
only adjust the predictive variance without affecting the mean. Consequently, 
optimizing $\Omega$ by maximizing the VI ELBO in Equation~\eqref{eq:opt_kl2} can lead to undesirable 
solutions where the predictive variance is set to zero. 

In this work, we propose the use of an \(\alpha\)-divergence objective (with \(\alpha = 1\)) to surpass the above mentioned limitation; which allows to train directly with a negative log-likelihood (NLL) term together with the usual KL regularizer, without any extra considerations. In practice, this change is sufficient to avoid degenerate solutions and results in non-zero, well-behaved predictive variances. The use of \(\alpha\)-divergences for approximate inference has been widely explored 
\citep{hernandez2016black, bui2017, SantanaH22}, with findings indicating that 
values of \(\alpha \approx 0\) enhance predictive mean estimation, 
while \(\alpha \approx 1\) improve predictive distributions, reflected in higher 
test log-likelihood performance. Thus, instead of minimizing \(\mathrm{KL}(q(f)|p(f|\mathbf{y}))\), our
objective is changed using a generalized view of VI \citep{knoblauch2022optimization} to minimize the \(\alpha\)-divergence between \(p(f|\mathbf{y})\) and \(q(f)\),
in an approximate way, for $\alpha \approx 1.0$ \citep{liG17}.  This can be 
achieved by changing the data-dependent term of the loss to the Black-Box \(\alpha\) (BB-\(\alpha\)) form:
\begin{equation}
	\label{eq:loss-alpha}
\mathcal{L}_{\alpha}(\Gamma)
=
\frac{1}{\alpha}\log \mathbb{E}_{q(f)} \left[ p(\mathbf{y}|f)^{\alpha}\right]
-
\mathrm{KL}\big(q \big| p\big)\,,
\qquad \alpha \in (0,1],
\end{equation}
with \(\Gamma =\{\bm{A}, \mathbf{Z}_\beta, \Omega, \sigma^2\}\), where the expectation is now inside the logarithm.
This expression makes the trade-off controlled by \(\alpha\) explicit: as \(\alpha \to 0\) we recover the standard
VI data term \(\mathbb{E}_{q(f)}[\log p(\mathbf{y}|f)]\), whereas setting \(\alpha = 1\) yields
\(\log \mathbb{E}_{q(f)}[p(\mathbf{y}|f)]\), which is the objective we adopt in practice. It is worth mentioning that in regression settings with Gaussian noise with variance \(\sigma^2\) the data dependent VI term (\(\alpha = 0\)) becomes:
\begin{equation*}
\sum_{i=1}^N -\frac{\log(2\pi\sigma^2)}{2} - \frac{(y_i - g(\mathbf{x}_i))^2}{2\sigma^2} 
	- \frac{K^\star(\mathbf{x}_i, \mathbf{x}_i)}{2\sigma^2}
\end{equation*}
where \((y_i - g(\mathbf{x}_i))^2\) is constant. This makes posterior covariances $K^\star(\mathbf{x}_i, \mathbf{x}_i)$ tend to $0$. See Appendix~\ref{subsec:ablation_alpha} for an ablation study on the effect of \(\alpha\).

\paragraph{Mini-batch Optimization} The objective in Equation~\eqref{eq:loss-alpha} supports mini-batch optimization with a cost in \(\mathcal{O}(M_\beta^3 + |\mathcal{B}|M_\beta^2)\):
\begin{equation}
	\mathcal{L}(\Gamma) \approx \frac{N}{|\mathcal{B}|} \sum_{b\, \in\, \mathcal{B}} \log \mathbb{E}_{q(f)} 
	\left[p(y_b|f)\right] -\mathrm{KL}\left(q| p\right)\,,
\end{equation}
where \(\mathcal{B}\) is a mini-batch of points. The expectation can be computed in closed form 
in regression. In classification, we use the approximation that is available via the softmax method described
in \cite{daxberger2021}.

\paragraph{Implementation} For clarity, we provide closed-form computation of the loss function in regression and classification settings in Appendix~\ref{app:closed-form}; and pseudocode for FMGP training and prediction in Appendix~\ref{app:code}.

\subsection{Limitations}
\label{sec:limitations}

The use of fixed-mean Gaussian processes for deep learning is limited by three factors: 
\begin{enumerate}[itemsep=-0.ex, partopsep=1ex, parsep=1ex]
	\item Computing the predictive distribution at each training iteration involves inverting \(\tilde{\bm{A}}^{-1} + \bm{K}_{\mathbf{Z}_\beta}\), 
	with cubic cost in the number of inducing points $M_{\beta}$. However, as shown in the experiments, this number can be set to a very low value, such as \(20\), even for classification tasks with a thousand classes.
	\item FMGPs require additional optimization steps compared to other \emph{post-hoc} approximations. However, other methods often rely    on visiting \emph{every training point} to compute specific updates. As a result, FMGP training can be faster in large datasets featuring millions of instances such as ImageNet.
    \item The construction of FMGP requires choosing a (parametric) kernel. This is both an advantage, 
	  as it may better capture the underlying data patterns in the modelling process, and a disadvantage, as it may be difficult to efficiently use an effective kernel in some tasks, such as image classification.
\end{enumerate}

\section{Related Work}\label{sec:related_work}
Due to its \emph{post-hoc} nature, FMGP is closely related to the Linearized Laplace Approximation (LLA) for deep learning. The Laplace Approximation (LA) \citep{mackay1992evidence} models the DNN posterior in parameter space by a Gaussian centered at the MAP estimate with covariance given by the inverse posterior Hessian. Scalability is improved by replacing the Hessian with a Generalized Gauss--Newton (GGN) approximation, which is equivalent to linearizing the network. This linearization at prediction time yields the \emph{post-hoc} LLA method \citep{ritter2018scalable}, which also mitigates some underfitting issues of LA \citep{lawrence2001variational}. In practice, however, the GGN matrix remains intractable and is typically approximated (e.g., via KFAC). Recent work has further improved LLA scalability and accuracy through Nystr\"om methods \citep{deng2022accelerated}, variational approaches \citep{ortega2024variational, scannell2024function}, sample-based approximations \citep{antoran2023sampling}, subrogate kernel learning \citep{ortega2026scalable} and quadratic approximations \citep{jimenez2026improving}. 

Among LLA-based methods, FMGP is most closely related to Variational LLA (VaLLA) \citep{ortega2024variational}. VaLLA interprets LLA as an induced GP \citep{khan2019approximate,immer2021improving} (enabled by the linearization) and uses a variational sparse GP to approximate posterior variances. When the Neural Tangent Kernel (NTK) is used, i.e.\ $\phi_{\mathbf{x}}=\partial g(\mathbf{x})/\partial\bm{\theta}$, VaLLA can be recovered as a special case of FMGP under additional assumptions. The main differences are: (i) FMGP does not rely on the NTK, enabling task-adapted kernels and avoiding repeated Jacobian computations, which can make VaLLA prediction costly on large-scale problems (e.g.\ ImageNet). In addition, FMGP's kernel flexibility can improve both predictive quality and efficiency. (ii) FMGP does not require an LLA approximation, so optimality of the GGN-based LLA is not central to the framework. (iii) The NTK is not guaranteed to be universal, and VaLLA effectively assumes $g(\cdot)\in\mathcal{H}$, which may fail in practice. (iv) VaLLA does not incorporate an explicit regularization mechanism to prevent overfitting, typically relying on early stopping and a validation set. In summary, VaLLA performs variational inference on the GP induced by LLA, whereas FMGP uses VI to select an optimal distribution within a family of Gaussian measures with fixed mean.

In \cite{deng2022accelerated}, the authors propose a Nystr\"om approximation of the GGN Hessian approximation of LLA using $M\ll N$ points chosen at random from the training set. The method, called ELLA, has cost $\mathcal{O}(NM^3)$. ELLA also requires computing the costly Jacobian vectors required in VaLLA, but does not need their gradients. Unlike VaLLA, the Nystr\"om approximation needs to visit each instance in the training set. However, as stated in \cite{deng2022accelerated}, ELLA suffers from over-fitting. Again, an early-stopping strategy using a validation set is proposed to alleviate it. In this case, ELLA only considers a subset of the training data. ELLA does not allow for hyper-parameter optimization, unlike VaLLA. The prior variance $\sigma_0^2$ must be tuned using grid search and a validation set, increasing the required training time significantly. 

Samples from the GP posterior corresponding to LLA can be generated via stochastic optimization, avoiding kernel matrix inversion and the resulting $\mathcal{O}(N^3)$ cost \citep{lin2023sampling,antoran2023sampling}. However, this approach does not provide a log-marginal likelihood for hyperparameter optimization. To address this, \cite{antoran2023sampling} propose an EM-style procedure that alternates between sample generation (E-step) and hyperparameter optimization (M-step). This substantially increases training cost, as generating a single sample can be as expensive as training the original DNN on the full dataset. Moreover, their study is limited to classification, and empirical evidence indicates that VaLLA is both faster and yields better performance.

Another GP-based approach for obtaining prediction uncertainty in the context of DNNs is the Spectral-normalized Neural Gaussian Process (SNGP) \citep{liu2023simple}, which replaces the last layer of the DNN with a GP. SNGP allows to either (i) fine-tune a pre-trained DNN model, or (ii) train a full DNN model from scratch. We compare results with the former in our experiments. However, we have observed that replacing the last layer with a GP does not keep the predictive mean as the output of the pre-trained DNN and often results in a drop in prediction performance. This is also observed in \cite{liu2023simple}. \citet{bergna2025post} uses a Gaussian process as activation function in the first layer, and \citet{calvo2026richer} leverages a projection of the Neural Tangent Kernel features.

Another simple option to transform a pre-trained DNN model to a Bayesian one is to consider a mean-field VI approximation of the DNN posterior where the means are initialized to the pre-trained optimal solution weights and kept fixed. This is known as \emph{mean-field VI fine-tuning} \citep{deng2023bayesadapter} and, as demonstrated in our experiments, it can achieve good results in terms of both prediction performance and uncertainty estimation. However, this method demands full training of the variance of each weight, which can be very costly and may require several training epochs. Furthermore, this method provides no closed-form predictive distribution. It relies on generating Monte Carlo samples to make predictions. As a result, further approximations must be considered to reduce the training time, such as \emph{Flipout Trick}~\citep{wen2018flipout}. Even though these techniques successfully reduce the training time, the required Monte Carlo samples significantly increase prediction time.

Finally, FMGP extends the work of \cite{cheng2017variational} by showing theoretically that one can fix the predictive mean to be equal to the output of a DNN,
if a universal kernel is considered. However, this change may lead to practical problems for hyper-parameter estimation. Specifically, there are degenerate solutions 
for the model's hyperparameters where the predictive variance collapses to zero. To address this issue, FMGP introduces a specific regularization technique that 
efficiently estimates the model's hyperparameters.

\section{Experiments}

We compare our proposed method, FMGP, with other methods including: last-layer LLA with and without KFAC approximation, ELLA~\citep{deng2022accelerated}, VaLLA~\citep{ortega2024variational}, a mean-field VI fine-tuning approach \citep{deng2023bayesadapter} and SNGP \citep{liu2023simple}. FMGP and VaLLA use \(100\) inducing points, as in \cite{ortega2024variational}. See Appendix~\ref{subsec:ablation_M} for an ablation study on the number of inducing points. ELLA employs \(2\,000\) random points and \(20\) random features as in \cite{deng2023bayesadapter}. All the timed experiments are executed on a NVIDIA A100 graphic card. Finally, an implementation of FMGP and all other methods is publicly available at \url{https://github.com/Ludvins/BayesiPy}.

\subsection{Synthetic Experiment}
\label{seq:synthetic_exp}

The experiment in Figure~\ref{fig:intro} illustrates the predictive distributions of commonly used Bayesian approaches on a heteroscedastic synthetic 1-dimensional dataset. It compares the predictive distribution of FMGPs against other methods, including: the linearized Laplace approximation (LLA) with hyper-parameters optimized to maximize the marginal likelihood; mean-field VI (MFVI) fine-tuning of the pre-trained model with Gaussian noise optimized on the training data; a zero mean prior GP with a squared exponential kernel, with hyper-parameters that maximize the marginal likelihood (GP Zero); the same GP with the pretrained model as prior mean (GP); and, Hamiltonian Monte Carlo (HMC) using a uniform prior for the variance of the Gaussian noise and the Gaussian prior over the DNN's weights.

Even though, HMC's predictions as usually understood as the \textit{gold standard} for assessing the predictive variances of other methods, in this setting it is unable of fitting the heteroscedastic noise. Furthermore, it does not scale to large problems. Figure~\ref{fig:intro} shows that MFVI gives predictive variances that are mainly constant in the missing data gap, while LLA tends to overestimate it by interpolating between data clusters. On the other hand, FMGP produces predictive variances comparable to those of HMC, while correctly modeling the noise nature.

The GP's predictive mean does not align with the DNN output and shifts the pre-trained solution, which is expected to worsen the resulting predictive performance. Moreover, the GP does not scale well to large problems. By contrast, FMGP not only produces predictive similar to those of HMC but also retains the predictive mean equal to the DNN's output, which is expected to result in improved prediction accuracy.

\subsection{Regression Problems}\label{sec:regression}

As part of the experimental evaluation, we consider three different large regression datasets:
\begin{enumerate}
    \item  The \textit{Year} dataset \citep{year_prediction_msd_203} with $515\,345$ instances and $90$ features. The data is divided as: the first $400\,000$ instances as train subset and the following $63\,715$ for validation. The rest of instances are taken for the test set.
    
    \item The \textit{US flight delay (Airline)} dataset \citep{dutordoir2020sparse}. Following \cite{ortega2023deep}, we use the first $600\,000$ instances for training, the following $100\,000$ instances for validation and the next $100\,000$ for testing. Here, $8$ features are considered: \textit{month}, \textit{day of the month}, \textit{day of the week}, \textit{plane age}, \textit{air time}, \textit{distance}, \textit{arrival time} and \textit{departure time}.
    
    \item  The \textit{Taxi dataset}, with data recorded on January, 2023. For this dataset, $9$ attributes are considered: \textit{time of day}, \textit{day of week}, \textit{day of month}, \textit{month}, \textit{PULocationID}, \textit{DOLocationID}, \textit{distance} and \textit{duration}; while the predictive variable is the \textit{price}. Following \citet{ortega2024variational}, we filter trips shorter than 10 seconds and larger than 5 hours, resulting in $3\,050\,311$ million instances. The first \(80\%\) is used as train data, the next \(10\%\) as validation data,  and the last \(10\%\) as testing data. 
\end{enumerate}
In all experiments, a pre-trained 3-layer DNN with 200 units with \emph{tanh} activations is employed. ELLA is trained without \textit{early-stopping} as over-fitting is not observed. Hyper-parameters are 
chosen using a grid search and the validation set. FMGP employs the squared-exponential kernel. MAP results are obtained by learning the optimal Gaussian 
noise using a validation set. A last-layer Kronecker approximation is used for LLA. 

\begin{figure}[t]
    \centering
    \includegraphics[width=1\linewidth]{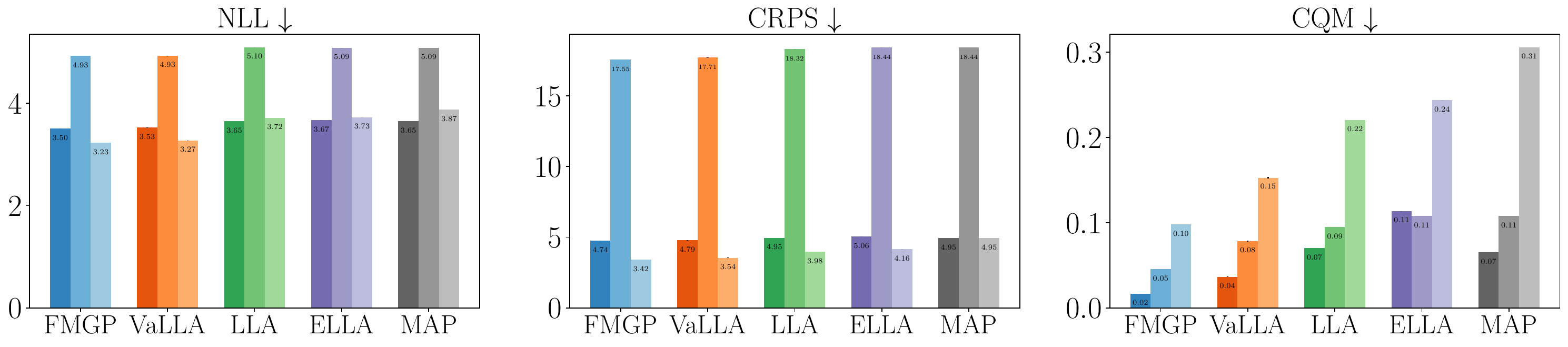}
    \caption{Results obtained in regression problems for different \emph{post-hoc} methods. Triple bars are shown corresponding to Year, Airline and Taxi datasets, from left to right. 
	MAP uncertainty is obtained using Gaussian noise optimized using a validation set. We report average results across \(5\) different repetitions using different random 
	seeds. Error-bars are shown but they are negligible in most cases.}
    \label{fig:regression}
\end{figure}

Figure~\ref{fig:regression} shows average results for each method over 5 different random seeds. We measure the quality of 
the predictive distribution in terms of the negative log likelihood (NLL), the continuous 
ranked probability score (CRPS) \citep{gneiting2007strictly} and a centered quantile metric (CQM) \citep{ortega2024variational}. 
Intuitively, CRPS can be understood as a generalization of the mean absolute error to predictive distributions. CQM measures 
the \emph{difference} between the models quantiles and the data quantiles under the same predictive mean, which is always the 
case here for each method. CQM is like a generalization of expected calibration error for regression problems. It is defined as:
\begin{align*}
	\text{CQM} = \int_0^1 \ \Big|\mathbb{P}_{(\mathbf{x}^\star, y^\star)}\left[ y^\star \in I(\mathbf{x}^\star, \alpha) \right] - \alpha\Big| \ d\alpha\,,
\end{align*}
where \(I(\mathbf{x}, \alpha)=(\lambda(-\alpha), \lambda(\alpha))\), $\lambda(\alpha) = \Phi_{\mu(\mathbf{x}),\sigma^2(\mathbf{x})}^{-1}(\tfrac{1 + \alpha}{2})$ and \(\Phi_{\mu(\mathbf{x}),\sigma^2(\mathbf{x})}\) is the CDF of a Gaussian with mean $\mu(\mathbf{x})$ and variance $\sigma^2(\mathbf{x})$, specified by the predictive distribution.

Figure~\ref{fig:regression} shows that FMGP performs best according to all three metrics (the lower the better), where the biggest 
difference is obtained in terms of CQM. As a result, we can argue that FMGP provides better uncertainty estimates (in terms of NLL) 
and calibration (both in terms of CRPS and CQM) compared to state-of-the-art LLA variants in regression settings.

\subsection{CIFAR10 Dataset and ResNet Architectures}

We perform experiments with various ResNet architectures \citep{he2016deep} on the CIFAR10 dataset \citep{cifar10}. 
To facilitate reproducibility, the considered pre-trained models are publicly available and accessible at \url{https://github.com/chenyaofo/pytorch-cifar-models}. The considered models include ResNet20 (\(272\, 474\) parameters), ResNet32 (\(466\, 906\) parameters), ResNet44 (\(661\, 338\) parameters) and ResNet56 (\(855\, 770\) parameters). Following~\citet{deng2022accelerated} and \citet{ortega2024variational}, ELLA and VaLLA use as validation set a data-augmented subset of \(5\, 000\) training points from the train set.
This validation set is obtained by performing random image crops of the training images of sizes in \([0.5, 1]\).

In multi-class classification problems, the kernel used in FMGP should model dependencies among the different DNN outputs, one per each class label. Therefore, we employ the following simple kernel in FMGP in that setting:
\begin{equation}
	K((\mathbf{x},c),(\mathbf{x}',c')) = 
	B_{c,c'} \times K_{RBF}(\mathbf{x}, \mathbf{x}') \times (\psi(\mathbf{x})^T \psi(\mathbf{x}') + \delta_{\mathbf{x}=\mathbf{x}'})\,,
\end{equation}
which includes a p.s. matrix \(\bm{B}\in \mathbb{R}^{C \times C}\) to model output dependencies, 
a squared exponential kernel in the input space, and a linear kernel plus noise in the high-level features \(\psi(\cdot)\), that correspond to the output of the pre-trained model up to the second-to-last layer. The trainable hyperparameters are the squared exponential amplitude and length scales of the RBF kernel (one per input feature), along with the matrix \(\bm{B}\), parameterized by its Cholesky decomposition. This simple kernel gives good results in our experiments. More sophisticated kernels are possible, potentially leading to even better results. The inducing points are randomly assigned to a class label.

\begin{figure}[t]
    \centering
    \includegraphics[width=0.99\linewidth]{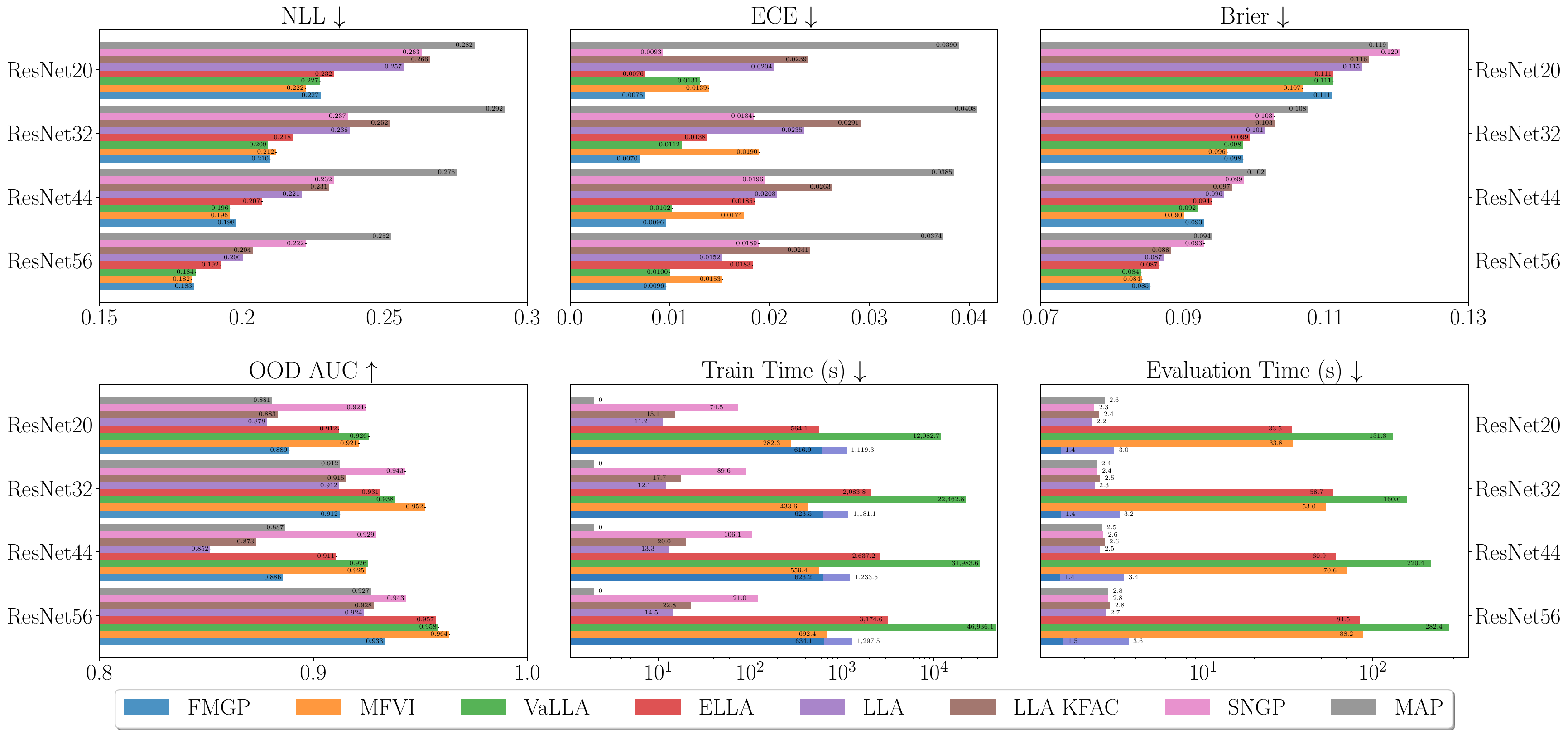}
    \caption{Test results obtained in CIFAR10 for different pre-trained ResNet architectures. LLA employs last-layer approximation. Out-of-distribution AUC is computed on a binary classification task discriminating between CIFAR10 and SVHN data instances. For this, we use the entropy of the predictive distribution. We report average results across \(5\) different repetitions using different random seeds. Error bars are shown but they are negligible in most cases.}
    \label{fig:cifar10}
\end{figure}

Figure~\ref{fig:cifar10} shows the negative log-likelihood (NLL), expected calibration error (ECE), and Brier score of each method. Furthermore, we also report the out-of-distribution AUC of each method 
in a binary classification problem with the SVHN dataset as the out-of-distribution data \citep{netzer2011reading}. In each method, we use predictive entropy as the decision function for classification between in and out-of-distribution. The training and evaluation times for each method are also reported. Recent work \citet{mucsanyi2024benchmarking} shows how different uncertainty quantification metrics tend to \emph{cluster} and the importance of measuring prediction uncertainty using as many as possible. Accuracy is not shown here as most methods barely change the pre-trained DNN accuracy. Notwithstanding, it is worth mentioning that SNGP tends to lower the accuracy of the model, as shown in \cite{liu2023simple}, while MFVI tends to increase it slightly, as noted in \cite{deng2022accelerated} and \cite{ortega2024variational}.

\begin{figure}[t]
    \centering
    \includegraphics[width=1\linewidth]{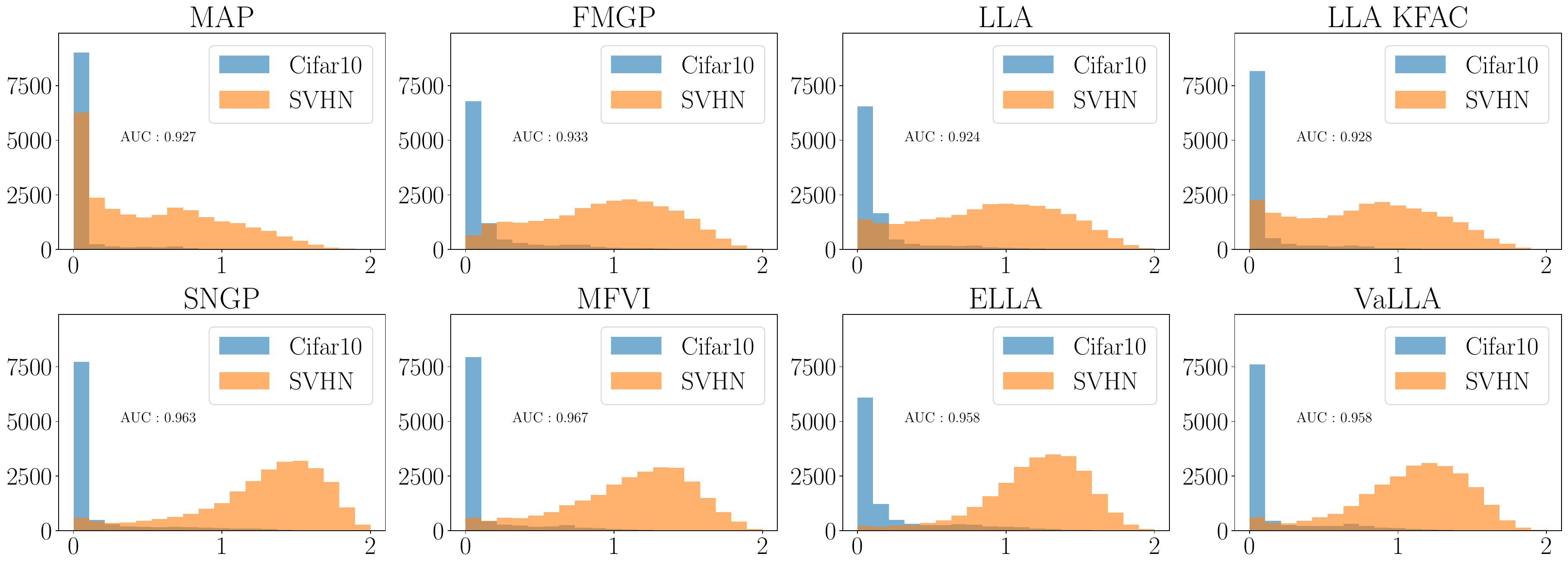}
    \caption{Histograms of the entropy of the predictive distribution of each method. We plot histograms for each class label, across \(5\) different repetitions using different random seeds.  The class labels are CIFAR10 (in-distribution) and SVHN (out-of-distribution) instances. We consider the ResNet56 architecture. We also show the average AUC of each method. }
    \label{fig_cifar10ood}
\end{figure}

Figure~\ref{fig:cifar10} shows that FMGP, MFVI, VaLLA and ELLA provide the highest performance in terms 
of NLL and Brier scores (the lower the better). However, in terms of ECE (also the lower the better), 
SNGP, VaLLA and FMGP provide better-calibrated uncertainties. As a result, FMGP and VaLLA seem to 
provide better uncertainty quantification with better-calibrated predictive distributions. 
However, for out-of-distribution detection, the best AUC is obtained by MFVI, ELLA, VaLLA and SNGP. 
Figure~\ref{fig_cifar10ood} shows histograms of the entropy of the predictive distribution of each
method for each type of test data (in and out-of-distribution). We believe the poor results of FMGP
in this task are due to the kernel choice. More sophisticated kernels may improve FMGP's results in this setting as well.

\begin{figure}[t]
    \centering
    \includegraphics[width=0.99\linewidth]{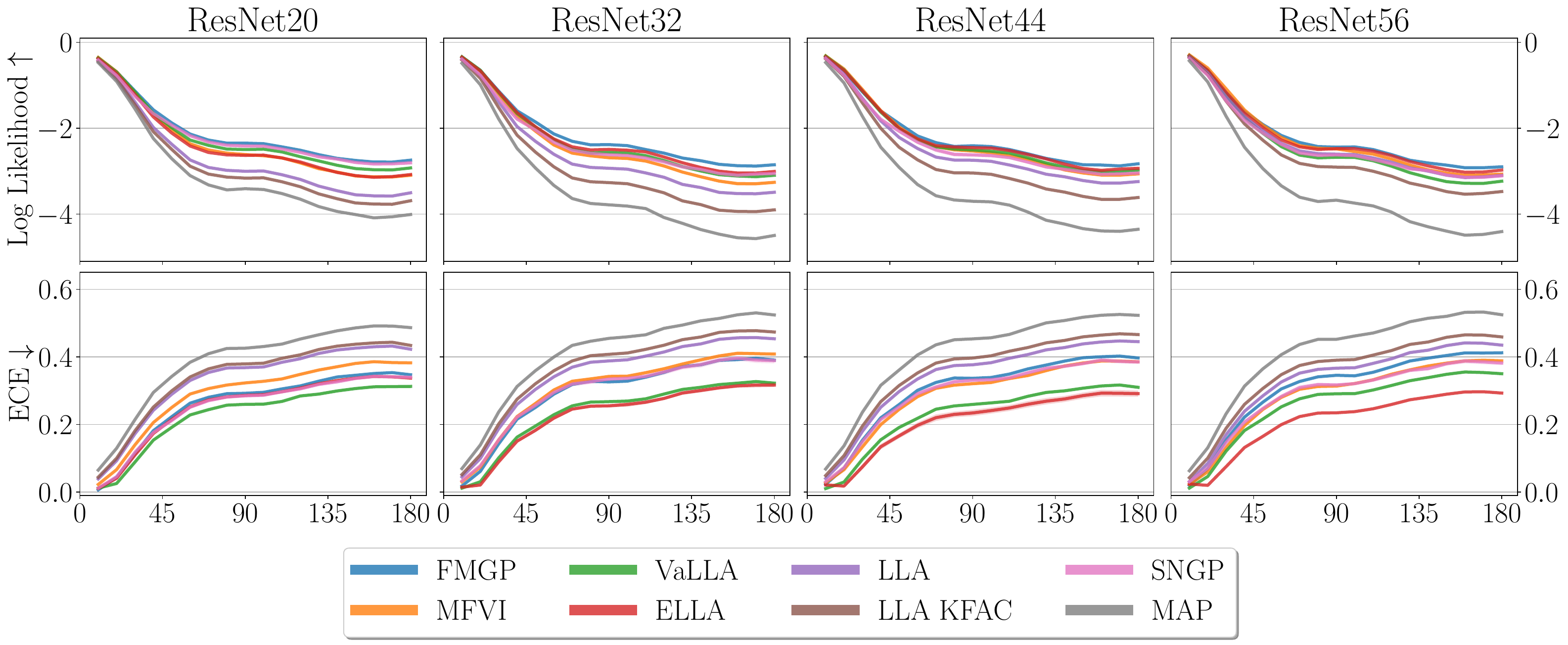}
    \caption{Robustness results obtained in rotated CIFAR10 for different pre-trained ResNet architectures. The x-axis corresponds to the degree of the rotations, from \(0^{\circ}\) to \(180^{\circ}\). LLA employs last-layer approximation.}
    \label{fig:cifar10_rotations}
\end{figure}

Regarding training time, Figure~\ref{fig:cifar10} shows that last-layer LLA approaches are the fastest to train, with VaLLA being the slowest method. At prediction time, SNGP, last-layer LLA and FMGP are quite similar to the pre-trained model. By contrast, VaLLA, ELLA and MFVI take larger prediction times. In VaLLA and ELLA this is due to the computation of the Jacobians, while in MFVI this due to Monte-Carlo sampling. Since FMGP is agnostic of the pre-trained model architecture, it only uses the DNN's predictions. Therefore, we also pre-computed all the model outputs and used them directly when training FMGP and making predictions using this model. As a result, a second bar is shown for FMGP indicating the training and evaluation time when pre-computing the outputs for both training and evaluation sets. In such a setting, the speed-up of FMGP is approximately \(\times1.5\) for training time and \(\times2.2\) for evaluation time.

Regarding predictive robustness, in Figure~\ref{fig:cifar10_rotations} we show the NLL and ECE of each method on rotated images of the CIFAR10 test set, as in \cite{ortega2024variational}. These results indicate that FMGP is the most robust method in terms of NLL, while it lies around the middle ground in terms of ECE. ELLA and VaLLA achieve the best results in this regard.

\begin{figure}[t]
    \centering
    \includegraphics[width=0.9\linewidth]{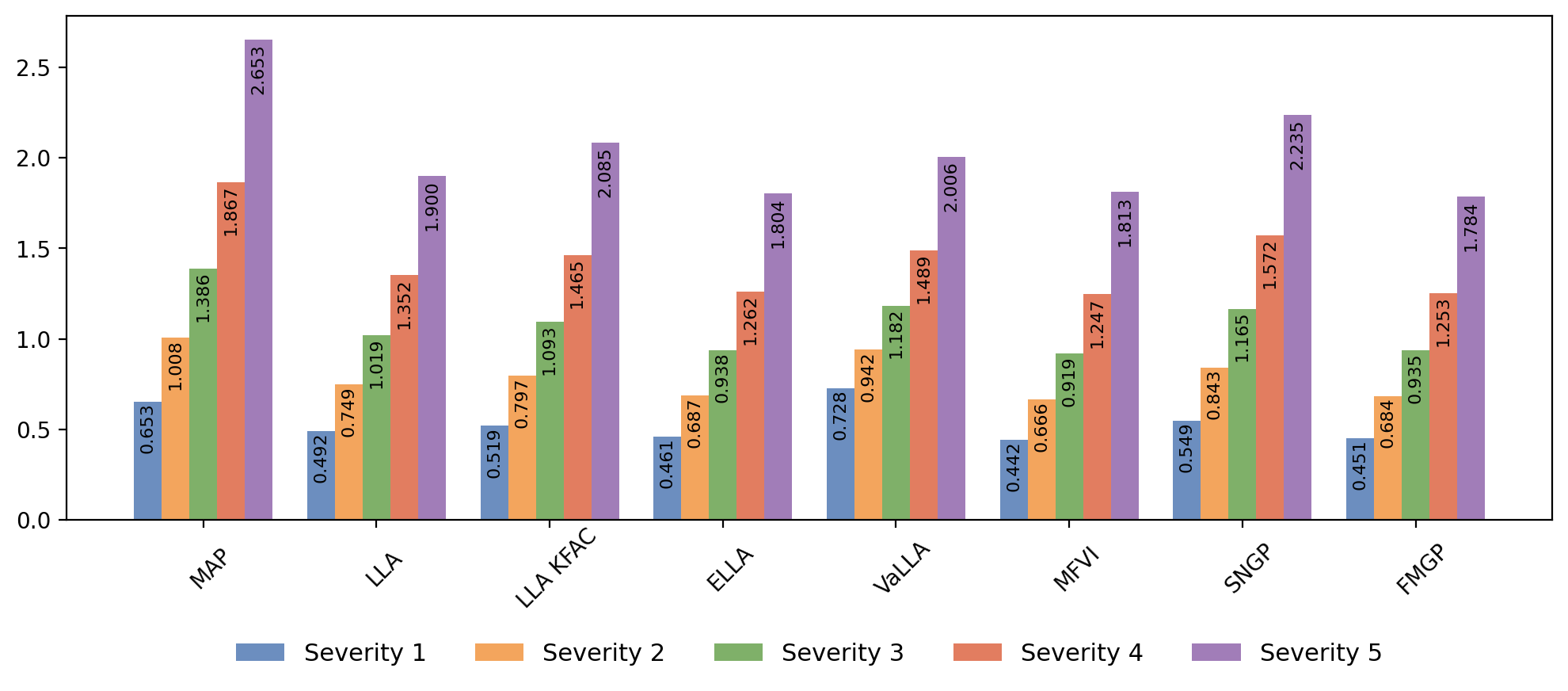}
    \caption{Robustness results (NLL) obtained in corrupted CIFAR10 ResNet56 and different post-hoc methods. LLA employs last-layer approximation.}
    \label{fig:cifar10_corruptions}
\end{figure}

Furthermore, Figure~\ref{fig:cifar10_corruptions} summarizes robustness on corrupted CIFAR10 (ResNet56) by reporting the average NLL across corruption types for increasing severity. As expected, NLL grows monotonically with severity for all methods, with the MAP baseline degrading the most (reaching $\approx 2.65$ at severity 5). Among post-hoc approaches, the strongest robustness is achieved by MFVI, ELLA, and FMGP, which consistently yield the lowest (or near-lowest) NLL across severities. In particular, MFVI is best for mild-to-moderate corruptions (severities 1--4), while FMGP becomes the top performer under the hardest setting (severity 5 $\approx 1.78$). Overall, these results suggest that function-space uncertainty methods provide more resilient predictive distributions under distribution shift than other methods, especially at high corruption levels.

\subsection{ImageNet Dataset and Extra ResNet Architectures}

We perform experiments with more ResNet architectures \citep{he2016deep} on the ImageNet 1k dataset \citep{ILSVRC15}. This dataset has \(1\,000\) different classes and over 1 million data instances. As pre-trained models, we considered those from TorchVision \citep{torchvision2016} available at 
\url{https://pytorch.org/vision/main/models/resnet.html}. Specifically, the considered models are ResNet18 (\(11\,689\,512\) parameters), ResNet34 (\(21\,797\,672\) parameters), ResNet50 (\(25\,557\,032\) parameters), ResNet101 (\(44\,549\,160\) parameters) and ResNet152 (\(60\,192\,808\) parameters). Importantly, due to the size of the DNNs and dataset, many methods became infeasible in these experiments. Specifically, LLA cannot be used even with last-layer approximations due to memory limitations. Furthermore, Monte Carlo sampling for MFVI testing takes longer than 1 day for models larger than ResNet18. For this reason, MFVI is only tested on the ResNet18 architecture. SNGP is not evaluated as it requires a training time of several days on the smallest architecture.

Table~\ref{tab:imagenet} shows the results obtained for each method on each ResNet architecture. 
The best method is highlighted in red and the second-to-best method is highlighted in green. We observe that, overall, FMGP obtains the best performance (NLL and ECE) while remaining the second-to-best in terms of computational time, only behind the MAP solution for the bigger models. As an additional detail, ELLA's validation set is computed using the same data-augmentation strategy proposed in \cite{deng2022accelerated}. 

Our results for ELLA are slightly different from those reported in \cite{deng2022accelerated} since ELLA's performance highly depends on the particular data augmentation performed to create the validation set. Despite using the same hyperparameters for this step, using the current PyTorch versions leads to different results.

\begin{table}[t]
    \centering
        \caption{Performance metrics for different methods and their train and test times in seconds on the ImageNet dataset}
    \label{tab:imagenet}
    \scalebox{0.9}{
	\begin{tabular}{llr@{$\pm$}lr@{$\pm$}lr@{$\pm$}lr@{$\pm$}l}
        \hline
		\textbf{Model} & \textbf{Method} & \multicolumn{2}{c}{\textbf{NLL}} & \multicolumn{2}{c}{\textbf{ECE}} & \multicolumn{2}{c}{\textbf{Train Time}} & \multicolumn{2}{c}{\textbf{Test Time}} \\
        \hline
		\multirow{4}{*}{ResNet18}  & MAP & {\color{teal} \bf{1.247}}&  {\color{teal} \bf{0.000}} & 0.026 & 0.000 & {\color{purple}\bf{0.0}} & {\color{purple}\bf{0.0}} & 
		{\color{purple}\bf{5.058}} & {\color{purple}{\bf 0.029$\times 10^2$}} \\
		& ELLA & 1.248 & 0.000 & {\color{teal} \bf{0.025}} & {\color{teal}{0.000}} & {\color{teal}\bf{7.890}} & {\color{teal} \bf{0.275$\times 10^3$}} & 8.060 & 0.010$\times 10^2$ \\
		& FMGP & 1.248 & 0.001 & {\color{purple}\bf{0.015}} & {\color{purple}{\bf 0.001}} & 1.835 & 0.099$\times 10^4$ & {\color{teal}\bf{7.324}} &  {\color{teal}\bf {0.001$\times 10^2$}} \\
		& MFVI & {\color{purple}\bf {1.242}} & {\color{purple}{\bf 0.001}} & 0.040 & 0.000 & 7.602 & 0.032$\times 10^4$& 3.773 & 0.308$\times 10^4$ \\
        \hline
		\multirow{3}{*}{ResNet34}  & MAP & {\color{teal}\bf{1.081}} &{\color{teal}\bf{0.000}} & 0.035 & 0.000& {\color{purple}\bf{0.0}} &{\color{purple}{\bf 0.0}} & {\color{purple}\bf{5.088}} & {\color{purple}{\bf 0.004$\times 10^2$}} \\
		& ELLA & 1.082 & 0.000 & {\color{teal}\bf{0.034}}  & {\color{teal}\bf{0.000}} & {\color{teal}\bf{1.201}}&  {\color{teal}{\bf 0.373$\times 10^4$}} & 1.087 & 0.018$\times 10^3$ \\
		& FMGP & {\color{purple}\bf{1.077}} & {\color{purple}{\bf 0.000}} & {\color{purple}\bf{0.016}}&  {\color{purple}\bf 0.000} & 1.942 & 0.103$\times 10^4$ & {\color{teal}\bf{8.563}} & {\color{teal}\bf{ 0.011$\times 10^2$}} \\
        \hline
		\multirow{3}{*}{ResNet50} & MAP & {\color{teal}\bf{0.962}} & {\color{teal}\bf{0.000}} & 0.037 & 0.000 & {\color{purple}\bf{0.0}} & {\color{purple}\bf{0.0}} & {\color{purple}\bf{4.954}} &  {\color{purple}\bf{0.010}$\times 10^2$} \\
		& ELLA & {\color{teal}\bf{0.962}} & {\color{teal}\bf{0.000}} & {\color{teal}\bf{0.036}} & {\color{teal}\bf{0.000}} & 2.997 & 1.215$\times 10^4$ & 1.954 & 0.018$\times 10^3$ \\
		& FMGP & {\color{purple}\bf{0.958}} & {\color{purple}\bf{0.001}} & {\color{purple}\bf{0.018}} & {\color{purple}\bf{0.001}} & {\color{teal}\bf{2.543}} & {\color{teal}\bf{0.046$\times 10^4$}} & {\color{teal}\bf{1.100}} & {\color{teal}\bf{0.010$\times 10^3$}} \\
        \hline
		\multirow{3}{*}{ResNet101} & MAP & {\color{teal}\bf{0.912}} & {\color{teal}\bf{0.000}} & 0.049 & 0.000 & {\color{purple}\bf{0.0}} & {\color{purple}\bf{0.0}} & {\color{purple}\bf{5.059}} & {\color{purple}\bf{0.001$\times 10^2$}} \\
		& ELLA & 0.913 & 0.000 & {\color{teal}\bf{0.048}} & {\color{teal}\bf{0.000}} & 4.464 & 1.649$\times 10^4$ & 2.808 & 0.001$\times 10^3$ \\
		& FMGP & {\color{purple}\bf{0.900}} & {\color{purple}\bf{0.000}} & {\color{purple}\bf{0.030}} & {\color{purple}\bf{0.001}} & {\color{teal}\bf{2.654}} & {\color{teal}\bf{0.064$\times 10^4$}} & {\color{teal}\bf{1.134}} & {\color{teal}\bf{0.001$\times 10^3$}} \\
        \hline
		\multirow{3}{*}{ResNet152} & MAP & {\color{teal}\bf{0.876}} & {\color{teal}\bf{0.000}} & 0.050 & 0.000& {\color{purple}\bf{0.0}} & {\color{purple}\bf{0.0}} & {\color{purple}\bf{6.324}} & {\color{purple}\bf{0.004$\times 10^2$}} \\
		& ELLA & 0.877 & 0.000 & {\color{teal}\bf{0.048}} & {\color{teal}\bf{0.000}} & 6.820 & 0.526$\times 10^4$ & 3.877 & 0.007$\times 10^3$ \\
		& FMGP & {\color{purple}\bf{0.865}} & {\color{purple}\bf{0.001}} & {\color{purple}\bf{0.024}} & {\color{purple}\bf{0.001}} & {\color{teal}\bf{2.973}} & {\color{teal}\bf{0.069$\times 10^4$}} & {\color{teal}\bf{1.267}} & {\color{teal}\bf{0.002$\times 10^3$}} \\
        \hline
    \end{tabular}}   
\end{table}

\subsection{Protein Feature Prediction Dataset}

QM9 is a dataset which provides quantum chemical properties (at DFT level) for a 
relevant, consistent, and comprehensive chemical space of around \( 130\,000 \) small organic 
molecules \citep{ruddigkeit2012enumeration}. In this experiment, we train a small convolutional neural network with message passing following the Torch-Geometric~\citep{Fey/Lenssen/2019} tutorial available at
\url{https://github.com/pyg-team/pytorch_geometric/}. 
The model is trained to make predictions on the \emph{dipole moment} target. 

In this regression experiment, the input space consists of molecules' graphs and not the usual tabular data considered in supervised learning. Therefore, in each method evaluated, we considered the model up to the last two linear functions to be a feature embedding of the graphs and assumed the data live in such embedding space. We used the first \(10\, 000\) data instances for testing, the following \(10\, 000\) data instances for validation, and the rest \(110\, 000\) instances for training. Here, ELLA is also trained without \textit{early-stopping} and the hyperparameters are chosen using a grid search on the validation set. FMGP employs the squared-exponential kernel with hyperparameters including the amplitude parameter and one length scale per dimension. MAP results are obtained by estimating the Gaussian noise on the validation set.

The results obtained are displayed in Table~\ref{tab:qm9} for MAP, last-layer Kronecker LLA, ELLA and FMGP in terms of the negative log-likelihood (NLL) and CRPS. We report average results across \(5\) repetitions of the experiments. The best result is again highlighted in purple and the second-best result in teal. We observe that FMGP provides the best performance (the smaller the better) in terms of both NLL and CRPS among the considered methods.

\begin{table}[t]
\centering
\caption{Results on QM9 dipole moment prediction task}\label{tab:qm9}
\begin{tabular}{lr@{$\pm$}lr@{$\pm$}l|lr@{$\pm$}lr@{$\pm$}l}
    \toprule
    \textbf{Method} & \multicolumn{2}{c}{\textbf{NLL}} & \multicolumn{2}{c}{\textbf{CRPS}} & \textbf{Method} & \multicolumn{2}{c}{\textbf{NLL}} & \multicolumn{2}{c}{\textbf{CRPS}} \\
    \midrule
    MAP & -1.76 & 0.016 & 0.0221 & 0.00 &     ELLA & {\color{teal}\bf{-1.80}} & {\color{teal}\bf{0.013}}  & 0.0219 & 0.00 \\
    LLA & -1.78 & 0.021 & {\color{teal}\bf{0.0218}} & {\color{teal}\bf{0.00}} & FMGP & {\color{purple}\bf{-1.85}} & {\color{purple}\bf{0.017}} & {\color{purple}\bf{0.0216}} & {\color{purple}\bf{0.00}} \\
\bottomrule
\end{tabular}
\end{table}

\subsection{CLIP Classification and FMGP}

CLIP (Contrastive Language–Image Pre-Training, \citep{radford2021learning}) is a neural network trained on a variety of (image, text) pairs. It can be instructed in natural language to predict the most relevant text snippet, given an image, without directly optimizing for the task, similarly to the zero-shot capabilities of GPT-2 and GPT-3. CLIP matches the performance of the original ResNet50 on ImageNet ``zero-shot'' without using any of the original \(1.28\)M labeled examples, overcoming several major challenges in computer vision.

In this experiment, we use CLIP on CIFAR10 and CIFAR100 in a ``black-box'' setting, and show how FMGPs can enhance the predictive capabilities of this approach. More precisely, consider a dataset 
\(\mathcal{D} = \{(\mathbf{x}_i, y_i)\}_{i=1}^N\)\ where \(\mathbf{x}_i\) denotes an image and \(y_i \in \{1, \ldots, C\}\) its corresponding class label. Each image \(\mathbf{x}_i\) is encoded using 
CLIP’s image encoder \(f_{\text{img}}(\cdot)\), and each class \(c_j\) is represented by a textual description 
\(
t_j = \text{``this is a } \{\text{class}\}\text{''}\,,
\)
where $\emph{class}$ is the class label word description of the image.
The CLIP classifier predicts the class by comparing the cosine similarities between the image and text embeddings, given by \(f_{\text{text}}(\cdot)\):
\begin{equation}
	p(y_i = j \mid \mathbf{x}_i) = 
	\frac{\exp\big(\tau \, \langle f_{\text{img}}(\mathbf{x}_i), f_{\text{text}}(t_j) \rangle \big)}
	{\sum_{k=1}^C \exp\big(\tau \, \langle f_{\text{img}}(\mathbf{x}_i), f_{\text{text}}(t_k) \rangle \big)},
\label{eq:clip_pred}
\end{equation}
where \(\tau\) is a temperature parameter learned during CLIP's pre-training, and \(\langle \cdot, \cdot \rangle\) denotes the cosine similarity.  

In our setting, we consider that we have access to the CLIP classifier as a black-box method. That is, we do not have access to its 
parameters, and hence, we cannot  compute Jacobians of the network output with respect to a particular input. Moreover, we do not have access to CLIP's original training inputs. We simply assume that we are given 
the classifier as such. Consequently, Laplace approximations or other Bayesian posterior approximations cannot be directly applied here.
However, FMGPs can be placed on top of the CLIP predictive scores to enhance calibration and capture residual 
uncertainty on the CIFAR10 and CIFAR100 datasets. In this setting, the CLIP logits act as a fixed mean function, while the GP predictive variances models 
the residual correlations across the image manifold. Instead of the squared exponential kernel, here, we simply use the dot product kernel 
computed on the CLIP high-level features associated to each image.

The results on the test set obtained for CLIP and CLIP combined with FMGP are displayed 
in Table \ref{tab:clip_fmgp_results}. The table shows that integrating an FMGP on top of the CLIP classifier 
consistently improves accuracy and reduces negative log-likelihood on both datasets, indicating better predictive 
calibration. The gains in terms of ECE and BRIER score are marginal or detrimental, however. 
In any case, we believe this experiment high-lights the potential benefits of FMGP for uncertainty modeling, in cases where
there is no access to the parameters of the underlying classifier.

\begin{table}[t]
\centering
\caption{Performance of CLIP and CLIP + FMGP (dot product kernel) on CIFAR10 and CIFAR100. Arrows indicate desired direction of improvement.}
\label{tab:clip_fmgp_results}
\begin{tabular}{lcccccc}
\toprule
\textbf{Dataset} & \textbf{Model} & \textbf{NLL} $\downarrow$ & \textbf{ACC} $\uparrow$ & \textbf{ECE} $\downarrow$ & \textbf{BRIER} $\downarrow$ \\
\midrule
\multirow{2}{*}{CIFAR10} 
  & CLIP & \(0.351\) & \(0.8873\) & \(\color{purple}\textbf{0.0113}\) & \(0.1671\) \\
  & CLIP + FMGP & $\color{purple}\textbf{0.339}$ & $\color{purple}\textbf{0.8914}$ & $0.0388$ & $\color{purple}\textbf{0.1651}$ \\
\midrule
\multirow{2}{*}{CIFAR100} 
  & CLIP & $1.426$ & $0.6170$ & $\color{purple}\textbf{0.0260}$ & $\color{purple}\textbf{0.5068}$ \\
  & CLIP + FMGP & $\color{purple}\textbf{1.375}$ & $\color{purple}\textbf{0.6337}$ & $0.1071$ & $0.5080$ \\
\bottomrule
\end{tabular}
\end{table}

\section{Conclusions}

In this work, we have introduced a method called fixed-mean Gaussian Processes. FMGPs leverage a family of variational distributions derived from the dual formulation of sparse GPs. This family corresponds to GPs where the predictive mean is fixed to any continuous function when using a universal kernel. Specifically, we set the continuous function to be the output of a pre-trained DNN. In such case, FMGPs become a \emph{post-hoc} method that, given a pre-trained DNN, outputs error bars estimating the confidence of the DNN in its predictions. FMGPs are both easy and efficient to train.

As demonstrated in our experiments, FMGPs excel at computing error bars for pre-trained DNNs with a large number of parameters, across a wide variety of performance metrics, on extensive datasets, handling millions of training instances, parameters, and thousands of output dimensions. Furthermore, FMGPs are applicable to a broad range of problems, including regression and classification tasks, where stochastic optimization enables sub-linear training costs with respect to the number of training instances.

Compared to other \emph{post-hoc} state-of-the-art methods for uncertainty estimation, FMGPs provide robust predictive distributions with minimal evaluation time. This efficiency stems from FMGPs relying solely on the outputs of the pre-trained DNN, without depending on its architecture or requiring the computation of DNN Jacobians, unlike related Linearized Laplace Approximation methods.

The proposed method is also useful in scenarios in which a direct access to the parameters of the pre-trained DNN may not be available. In such a setting, typical LLA approximate methods cannot be
used to estimate predictive variances since they require expensive Jacobian computations. By contrast, FMGP has no problem being used in that setting and can efficiently estimate the prediction uncertainty.

\acks{
The authors acknowledge financial support
from the project
PID2022-139856NB-I00,
funded by MCIN\slash AEI\slash 10.13039\slash
501100011033\slash FEDER, UE;
from project IDEA-CM (TEC-2024\slash COM-89),
funded by the Autonomous Community of Madrid;
and from the ELLIS Unit Madrid.
The authors also acknowledge computational support
from the Centro de Computaci\'on Cient\'ifica-Universidad
Aut\'onoma de Madrid (CCC-UAM).
}

\bibliography{references}

\newpage

\appendix

\section{Closed-form formulas}\label{app:closed-form}

This appendix provides explicit expressions for the FMGP training objective in
regression and classification settings.
We use the Black-Box $\alpha$ (BB-$\alpha$) objective in Equation~\eqref{eq:loss-alpha},
and in practice we set $\alpha=1$, yielding the data term
$\log \mathbb{E}_{q(f)}[p(\mathbf y|f)]$.

\subsection{Mini-batch BB-$\alpha$ objective }
Let $B$ be a mini-batch of indices, $|B|$ its size, and $N$ the total number of
training points. For each $b\in B$ we denote by $q(f_b)$ the marginal of the
variational process at $\mathbf x_b$ (or at $(\mathbf x_b,c)$ in classification).
The mini-batch approximation of Equation~\eqref{eq:loss-alpha} is
\begin{equation}
\widehat{\mathcal L}_{\alpha}(\Gamma)
=
\frac{N}{|B|}
\sum_{b\in B}
\frac{1}{\alpha}
\log \mathbb E_{q(f_b)} \big[p(y_b|f_b)^{\alpha}\big]
-
\mathrm{KL}(q|p)\,,
\qquad \alpha\in(0,1].
\end{equation}

\subsection{Regression: Gaussian likelihood}
Assume homoscedastic Gaussian noise with variance $\sigma^2$:
\[
p(y_b|f_b)=\mathcal N(y_b|f_b,\sigma^2).
\]
Let $f_b\sim \mathcal N(\mu_b,v_b)$ be any scalar Gaussian marginal. Then the BB-$\alpha$ expectation has a closed form:
\begin{align*}
\mathbb E \left[p(y_b|f_b)^{\alpha}\right]
&=
(2\pi\sigma^2)^{-\alpha/2}
\sqrt{\frac{\sigma^2}{\sigma^2+\alpha v_b}}\;
\exp \left(
-\frac{\alpha (y_b-\mu_b)^2}{2(\sigma^2+\alpha v_b)}
\right),
\\
\frac{1}{\alpha}\log \mathbb E \left[p(y_b|f_b)^{\alpha}\right]
&=
-\frac{1}{2}\log(2\pi\sigma^2)
-\frac{1}{2\alpha}\log \Big(1+\alpha \frac{v_b}{\sigma^2}\Big)
-\frac{(y_b-\mu_b)^2}{2(\sigma^2+\alpha v_b)}.
\end{align*}
Setting $\alpha=1$ gives the particularly simple expression
\begin{equation*}
\log \mathbb E \left[p(y_b|f_b)\right]
=
\log \mathcal N \big(y_b|\mu_b,\,\sigma^2+v_b\big)
=
-\frac{1}{2}\log \big(2\pi(\sigma^2+v_b)\big)
-\frac{(y_b-\mu_b)^2}{2(\sigma^2+v_b)}.
\end{equation*}
In FMGP regression we plug $\mu_b=g(\mathbf x_b)$ for the fixed-mean term.

\subsection{Classification: softmax likelihood}
In multi-class classification, let $y_b\in\{1,\dots,C\}$ and
\[
p(y_b=c \mid \mathbf f_b)=\mathrm{softmax}(\mathbf f_b)_c
=
\frac{\exp(f_{b,c})}{\sum_{j=1}^C \exp(f_{b,j})}.
\]
For $\alpha=1$, the required data term is
\begin{equation}
\label{eq:softmax_dataterm}
\ell_b
=
\log \mathbb E_{\mathbf f_b\sim \mathcal N(\bm\mu_b,\bm V_b)}
\Big[\mathrm{softmax}(\mathbf f_b)_{y_b}\Big]\,.
\end{equation}
This Gaussian--softmax integral is not available in closed form, so we use either:

\subsubsection{Monte Carlo estimate}
Draw $S$ samples $\mathbf f_b^{(s)}\sim \mathcal N(\bm\mu_b,\bm V_b)$ and compute
\begin{equation*}
\widehat{\ell}_b^{\, \text{MC}}
=
\log\left(
\frac{1}{S}\sum_{s=1}^S \mathrm{softmax}(\mathbf f_b^{(s)})_{y_b}
\right).
\end{equation*}
For numerical stability one typically evaluates each softmax via log-sum-exp.

\subsubsection{Logit-scaling approximation}
To avoid Monte Carlo estimation, in classification a deterministic logit-scaling approximation exists \citep{daxberger2021}.
Given the Gaussian marginal
\[
\mathbf f_b \sim \mathcal N(\bm\mu_b,\bm V_b),\qquad y_b\in\{1,\dots,C\},
\]
we form \emph{scaled logits} $\widetilde{\bm\mu}_b\in\mathbb R^C$ by shrinking each class logit
according to its marginal variance (diagonal of $\bm V_b$):
\begin{equation*}
\label{eq:logit_scaling}
\widetilde{\mu}_{b,c}
=
\frac{\mu_{b,c}}{\sqrt{\,1+\frac{\pi}{8}\, [\bm V_b]_{c,c}\,}},
\qquad c=1,\dots,C.
\end{equation*}
We then approximate the predictive class probabilities as
\begin{equation}
\label{eq:logit_scaling_probs}
\widehat{\pi}_{b,c}
\;\approx\;
\mathrm{softmax}(\widetilde{\bm\mu}_b)^\alpha_c,
\qquad c=1,\dots,C.
\end{equation}
For the BB-$\alpha$ objective used in practice with $\alpha=1$, the data term
\[
\ell_b=\log \mathbb E_{\mathbf f_b}[\mathrm{softmax}(\mathbf f_b)_{y_b}^\alpha]
\]
is approximated by
\begin{equation}
\label{eq:logit_scaling_dataterm}
\widehat{\ell}_b
\;\approx\;
\log \widehat{\pi}_{b,y_b}
=
\log \mathrm{softmax}(\widetilde{\bm\mu}_b)_{y_b}^\alpha.
\end{equation}

\section{Pseudo-code}\label{app:code}

This section summarizes the FMGP procedure in algorithmic form, separating training and inference for clarity. 
Algorithm~\ref{alg:fmgp-train} describes the stochastic variational training phase.
Algorithm~\ref{alg:fmgp-predict} details test-time inference.
\begin{figure}[t]

\centering

% =========================
% Training Algorithm
% =========================
\begin{minipage}[t]{0.49\textwidth}
\vspace{0pt}
\begin{algorithm}[H]
\scriptsize
\setlength{\algomargin}{0.5em}
\SetAlgoNlRelativeSize{-1}
\SetAlgoInsideSkip{smallskip}
\caption{\scriptsize Training FMGP}
\label{alg:fmgp-train}
\DontPrintSemicolon
\SetKwInOut{KwIn}{Input}
\SetKwInOut{KwOut}{Output}

\KwIn{Dataset $\mathcal D=\{(\mathbf x_i,y_i)\}_{i=1}^N$, pretrained predictor $g(\cdot)$, kernel $K_\Omega$, inducing size $M_\beta$, batch size $|B|$, learning rate $\eta$}
\KwOut{Optimized parameters $\{\widetilde{\bm A},\mathbf Z_\beta,\Omega,\sigma^2\}$}

Initialize inducing locations $\mathbf Z_\beta$ (e.g.\ K-means)\;
Initialize $\widetilde{\bm A}\leftarrow\bm I_{M_\beta}$\;
Initialize kernel $\Omega$ and $\sigma^2$\;

\For{$t=1$ \KwTo $T$}{

  Sample mini-batch $B\subset\{1,\dots,N\}$\;

  Compute prior matrix:
  $\bm K_\beta \leftarrow K_\Omega(\mathbf Z_\beta,\mathbf Z_\beta)$\;
  Form covariance parameterization:
  \[
  \bm A \leftarrow -(\widetilde{\bm A}^{-1}+\bm K_\beta)^{-1}
  \]
  \ForEach{$b\in B$}{
      Compute marginal variance:
      \[
      v_b = K_\Omega(\mathbf x_b,\mathbf x_b)
      +K_\Omega(\mathbf x_b,\mathbf Z_\beta)
      \bm A
      K_\Omega(\mathbf Z_\beta,\mathbf x_b)
      \]

      Form predictive marginal:
      \[
      f_b \sim \mathcal N(g(\mathbf x_b), v_b)
      \]

      Evaluate expected log-likelihood:
      \[
      \ell_b \leftarrow
      \log \mathbb E_{f_b}[p(y_b|f_b)]
      \]
  }
  Compute KL term:
  \[
  \begin{aligned}
  \mathrm{KL}(q|p)
  &=
  \tfrac12
  \log\left|
  \bm I-\bm K_\beta(\bm A^{-1}+\bm K_\beta)^{-1}
  \right|\\
  &-
  \tfrac12\mathrm{tr}(\bm K_\beta\bm A)
  + C
  \end{aligned}
  \]
  Form stochastic objective:
  \[
  \widehat{\mathcal L}
  =
  \frac{N}{|B|}\sum_{b\in B}\ell_b
  -
  \mathrm{KL}(q|p)
  \]
  Update
  $\Gamma=\{\widetilde{\bm A},\mathbf Z_\beta,\Omega,\sigma^2\}$
  via Adam$(\eta)$\;
}

\Return optimized parameters\;
\end{algorithm}
\end{minipage}
\hfill
% =========================
% Prediction Algorithm
% =========================
\begin{minipage}[t]{0.49\textwidth}
\vspace{0pt}
\begin{algorithm}[H]
\scriptsize
\setlength{\algomargin}{0.5em}
\SetAlgoNlRelativeSize{-1}
\SetAlgoInsideSkip{smallskip}
\caption{\scriptsize Prediction with FMGP}
\label{alg:fmgp-predict}
\DontPrintSemicolon
\SetKwInOut{KwIn}{Input}
\SetKwInOut{KwOut}{Output}

\KwIn{Test input $\mathbf x_\star$, predictor $g(\cdot)$,
trained parameters $\{\widetilde{\bm A},\mathbf Z_\beta,\Omega,\sigma^2\}$}
\KwOut{Predictive mean $\bm\mu_\star$, variance $\mathbf v_\star$, predictive distribution}

Compute prior matrix:
$\bm K_\beta \leftarrow K_\Omega(\mathbf Z_\beta,\mathbf Z_\beta)$\;

Recover covariance parameter:
\[
\bm A \leftarrow
-(\widetilde{\bm A}^{-1}+\bm K_\beta)^{-1}
\]

Predictive mean:
\[
\bm\mu_\star \leftarrow g(\mathbf x_\star)
\]

Predictive latent variance:
\[
\mathbf v_\star =
K_\Omega(\mathbf x_\star,\mathbf x_\star)
+
K_\Omega(\mathbf x_\star,\mathbf Z_\beta)
\bm A
K_\Omega(\mathbf Z_\beta,\mathbf x_\star)
\]

\If{regression}{
  \[
  p(y_\star|\mathbf x_\star)
  =
  \mathcal N(
  \bm\mu_\star,
  \mathbf v_\star+\sigma^2)
  \]
}
\Else{
  \[
  p(y_\star=c|\mathbf x_\star)
  \approx
  \mathbb E_{\mathbf f_\star
  \sim\mathcal N(\bm\mu_\star,\mathbf v_\star)}
  [\mathrm{softmax}(\mathbf f_\star)_c]
  \]
}

\Return predictive quantities\;
\end{algorithm}
\end{minipage}

\end{figure}

\section{Ablation Studies}
\label{sec:ablation}

In this section we analyze the sensitivity of the proposed Fixed-Mean Gaussian Process (FMGP) model to two key design choices: (i) the $\alpha$ parameter of the BB-$\alpha$ divergence, and (ii) the number of inducing points $M$. All experiments are conducted on the \emph{Year} regression dataset described in Section~\ref{sec:regression}, using the same train/test splits and optimization protocol as in the main experiments.

Throughout this section we report test RMSE, negative log-likelihood (NLL), and continuous ranked probability score (CRPS). Lower values indicate better predictive performance.

% ---------------------------------------------------------------------
\subsection{$\alpha$-Divergences}\label{subsec:ablation_alpha}

We first study the impact of the $\alpha$ parameter in the BB-$\alpha$ objective. Recall that $\alpha$ interpolates between variational inference ($\alpha \to 0$) and expectation propagation ($\alpha = 1$), controlling the trade-off between mass-covering and mode-seeking behavior. Table~\ref{tab:ablation_alpha_year} reports the results for $\alpha \in \{0, 0.25, 0.5, 0.75, 1.0\}$ using $M=100$ inducing points. Several consistent trends emerge:

\paragraph{Predictive metrics.}
The probabilistic metrics (NLL and CRPS) improve monotonically as $\alpha$ increases. In particular, $\alpha=1.0$ achieves the best NLL and CRPS. This indicates that the choice of divergence primarily affects uncertainty calibration rather than mean predictions.

\paragraph{Noise variance behavior.}
For small values of $\alpha$ ($0$ and $0.25$), the model learns a non-negligible observation noise variance. As $\alpha$ increases, the learned noise variance decreases substantially and collapses to zero at $\alpha=1.0$. This is consistent with the fact that larger $\alpha$ values encourage sharper posterior approximations.

\paragraph{Kernel hyperparameters.}
The learned kernel amplitude and length-scales change dramatically across $\alpha$. For $\alpha \leq 0.25$, the amplitude collapses to zero and the model effectively reduces to the prior mean predictor. For $\alpha \geq 0.5$, the amplitude becomes large and the length-scales exhibit high variability, indicating that the covariance component actively compensates for the fixed mean.

To better understand this behavior, Table~\ref{tab:variance_decomposition_alpha_percent} decomposes the predictive variance into its prior contribution and the variance-correction term:
\[
	 \underbrace{K^{\star}_{\mathbf{Z}_\beta, \bm{A}}(\mathbf{x}, \mathbf{x}')}_{\mathrm{Posterior}}  = \underbrace{K(\mathbf{x}, \mathbf{x}')}_{\mathrm{Prior}} -  \underbrace{K(\mathbf{x}, \mathbf{Z}_\beta)(\tilde{\bm{A}}^{-1} + \bm{K}_\beta)^{-1}K(\mathbf{Z}_\beta, \mathbf{x}')}_{\mathrm{Correction}} \,.
\]
For $\alpha \in \{0, 0.25, 0.5\}$, the posterior variance is almost entirely explained by the prior term, meaning the learned correction is negligible. In contrast, for $\alpha \in \{0.75, 1.0\}$, the prior and correction terms contribute with comparable magnitude, indicating that the model learns a meaningful covariance adjustment around the fixed mean.

Overall, these results show that larger values of $\alpha$ encourage the model to depart from the prior and learn a non-trivial posterior covariance structure, leading to improved uncertainty estimates without affecting predictive accuracy in terms of RMSE.

% Appendix B -- Ablation: alpha-divergence (Year dataset, regression)
\begin{table}[t]
\centering
\small
\setlength{\tabcolsep}{6pt}
\begin{tabular}{c c c c c c}
\toprule
$\alpha$ & $M$ & Amp. & $\ell$ mean $\pm$ std & $\sigma^2$ & RMSE / NLL / CRPS \\
\midrule
0.00 & 100 & 0.0000  & $2.2239 \times 10^{3} \pm 2.3228 \times 10^{2}$  & 0.7200 & 9.3168 / 3.6508 / 4.9526 \\
0.25 & 100 & 0.0000  & $4.6336 \times 10^{3} \pm 2.7452 \times 10^{3}$  & 0.7226 & 9.3168 / 3.6508 / 4.9534 \\
0.50 & 100 & 2.5472  & $2.0825 \times 10^{7} \pm 1.2641 \times 10^{7}$  & 0.6865 & 9.3168 / 3.6315 / 4.9305 \\
0.75 & 100 & 15.7137 & $3.4197 \times 10^{7} \pm 2.7241 \times 10^{7}$  & 0.5031 & 9.3168 / 3.5738 / 4.8421 \\
1.00 & 100 & 13.6622 & $5.4784 \times 10^{5} \pm 2.2722 \times 10^{6}$  & 0.0000 & 9.3168 / 3.5085 / 4.7496 \\
\bottomrule
\end{tabular}
\caption{Ablation on the $\alpha$ parameter of BB-$\alpha$ divergences (Year dataset). Length-scales are reported in scientific notation. Lower is better for RMSE, NLL, and CRPS.}
\label{tab:ablation_alpha_year}
\end{table}

\begin{table}[t]
\centering
\small
\setlength{\tabcolsep}{6pt}
\begin{tabular}{c c c c}
\toprule
$\alpha$ & Prior (\%) & Posterior (\% of prior) & Var.\ Corr.\ (\% of prior) \\
\midrule
0.00 & 100.0 & 99.94 & 0.06 \\
0.25 & 100.0 & 99.9997 & 0.0003 \\
0.50 & 100.0 & 1.02 & 98.98 \\
0.75 & 100.0 & 1.40 & 98.60 \\
1.00 & 100.0 & 5.34 & 94.66 \\
\bottomrule
\end{tabular}
\caption{Predictive variance decomposition expressed as percentage of the prior variance for different values of $\alpha$ on the Year dataset.}
\label{tab:variance_decomposition_alpha_percent}
\end{table}

% ---------------------------------------------------------------------
\subsection{Number of inducing points}\label{subsec:ablation_M}

We next analyze the impact of the number of inducing points $M$. For this study, we fix $\alpha=1.0$, which yielded the best probabilistic performance in the previous subsection, and vary $M \in \{10, 50, 100, 200\}$. The results are shown in Table~\ref{tab:ablation_M_year}.

\paragraph{Predictive metrics.}
NLL and CRPS consistently improve as $M$ increases, showcasing that increasing the number of inducing locations leads to better calibrated predictive distributions.

\paragraph{Kernel hyperparameters.}
As $M$ increases, the learned length-scales decrease substantially and become more stable (lower standard deviation and narrower empirical range). This suggests that with more inducing points the model can represent finer-scale structure in the covariance function. The amplitude increases slightly with $M$, while the noise variance remains zero across all configurations.

\paragraph{Stability considerations.}
With very small $M$ (e.g., $M=10$), the length-scales become extremely large and highly variable, indicating that the model compensates for limited representational capacity by flattening the covariance. Increasing $M$ allows the posterior correction term to better approximate the full GP covariance, resulting in improved calibration and more stable hyperparameter estimates.

In summary, increasing the number of inducing points primarily improves uncertainty quantification. For the Year dataset, $M=100$ provides a favorable trade-off between performance and computational cost, with only marginal improvements observed at $M=200$.
% Appendix B -- Ablation: number of inducing points (Year dataset, regression)
\begin{table}[t]
\centering
\small
\setlength{\tabcolsep}{6pt}
\begin{tabular}{c c c c c c}
\toprule
$\alpha$ & $M$ & Amp. & $\ell$ mean $\pm$ std & $\sigma^2$ & RMSE / NLL / CRPS \\
\midrule
1.00 & 10  & 11.2417 & $9.1429 \times 10^{5} \pm 2.9772 \times 10^{6}$ & 0.0000 & 9.3168 / 3.5175 / 4.7622 \\
1.00 & 50  & 19.5535 & $1.5634 \times 10^{5} \pm 5.7386 \times 10^{5}$ & 0.0000 & 9.3168 / 3.4981 / 4.7363 \\
1.00 & 100 & 20.4026 & $4.1574 \times 10^{4} \pm 1.3661 \times 10^{5}$ & 0.0000 & 9.3168 / 3.4932 / 4.7327 \\
1.00 & 200 & 20.5211 & $1.6140 \times 10^{4} \pm 5.3422 \times 10^{4}$ & 0.0000 & 9.3168 / 3.4893 / 4.7291 \\
\bottomrule
\end{tabular}
\caption{Ablation on the number of inducing points $M$ for BB-$\alpha$ with $\alpha=1.0$ (Year dataset). Length-scales are reported in scientific notation.}
\label{tab:ablation_M_year}
\end{table}

\end{document}